%% file: main.tex
\documentclass[10pt,twocolumn,letterpaper]{article}

\usepackage[pagenumbers]{cvpr} 

\input{preamble}

\definecolor{cvprblue}{rgb}{0.21,0.49,0.74}
\usepackage[pagebackref,breaklinks,colorlinks,citecolor=cvprblue]{hyperref}

\usepackage[accsupp]{axessibility} 
\usepackage{bm}
\usepackage{mathtools}
\usepackage{amsmath}
\usepackage{pifont}
\usepackage{algorithm}
\usepackage{algpseudocode}
\usepackage{wrapfig}
\usepackage{booktabs}
\usepackage{pbox}
\usepackage{multicol}
\usepackage{multirow}
\usepackage{graphicx}
\usepackage{caption}
\usepackage{subcaption}
\usepackage{bbm}
\usepackage[table]{xcolor}
\usepackage{subcaption}

\newcommand*\dd{\mathop{}\!\mathrm{d}}
\newcommand{\e}[1]{{\small $#1$}}

\definecolor{rulecolor}{RGB}{0,71,171}
\definecolor{tableheadcolor}{RGB}{204,229,255}

\newcommand{\topline}{ %
    \arrayrulecolor{rulecolor}\specialrule{0.1em}{\abovetopsep}{0pt}%
    \arrayrulecolor{rulecolor}\specialrule{\lightrulewidth}{0pt}{0pt}%
    \arrayrulecolor{tableheadcolor}\specialrule{\aboverulesep}{0pt}{0pt}%
    \arrayrulecolor{rulecolor}
    }
\newcommand{\midtopline}{
    \arrayrulecolor{tableheadcolor}\specialrule{\aboverulesep}{0pt}{0pt}%
    \arrayrulecolor{rulecolor}\specialrule{\lightrulewidth}{0pt}{0pt}%
    \arrayrulecolor{white}\specialrule{\aboverulesep}{0pt}{0pt}%
    \arrayrulecolor{rulecolor}}
    \newcommand{\bottomline}{
    \arrayrulecolor{tableheadcolor}\specialrule{\aboverulesep}{0pt}{0pt}
    \arrayrulecolor{rulecolor}
    \specialrule{\heavyrulewidth}{0pt}{\belowbottomsep}
    \arrayrulecolor{rulecolor}\specialrule{\lightrulewidth}{0pt}{0pt}
    }
\newcolumntype{?}{!{\vrule width 1.4pt}}

\title{Correcting Diffusion Generation through Resampling}

\author{Yujian Liu\\
UC Santa Barbara\\
{\tt\small yujianliu@ucsb.edu}
\and
Yang Zhang\\
MIT-IBM Watson AI Lab\\
{\tt\small yang.zhang2@ibm.com}
\and
Tommi Jaakkola\\
MIT CSAIL\\
{\tt\small tommi@csail.mit.edu}
\and
Shiyu Chang\\
UC Santa Barbara\\
{\tt\small chang87@ucsb.edu}
}

\begin{document}
\let\oldtwocolumn\twocolumn
\renewcommand\twocolumn[1][]{%
    \oldtwocolumn[{#1}{
    \begin{center}
\vspace{-3mm}
\includegraphics[width=0.91\textwidth]{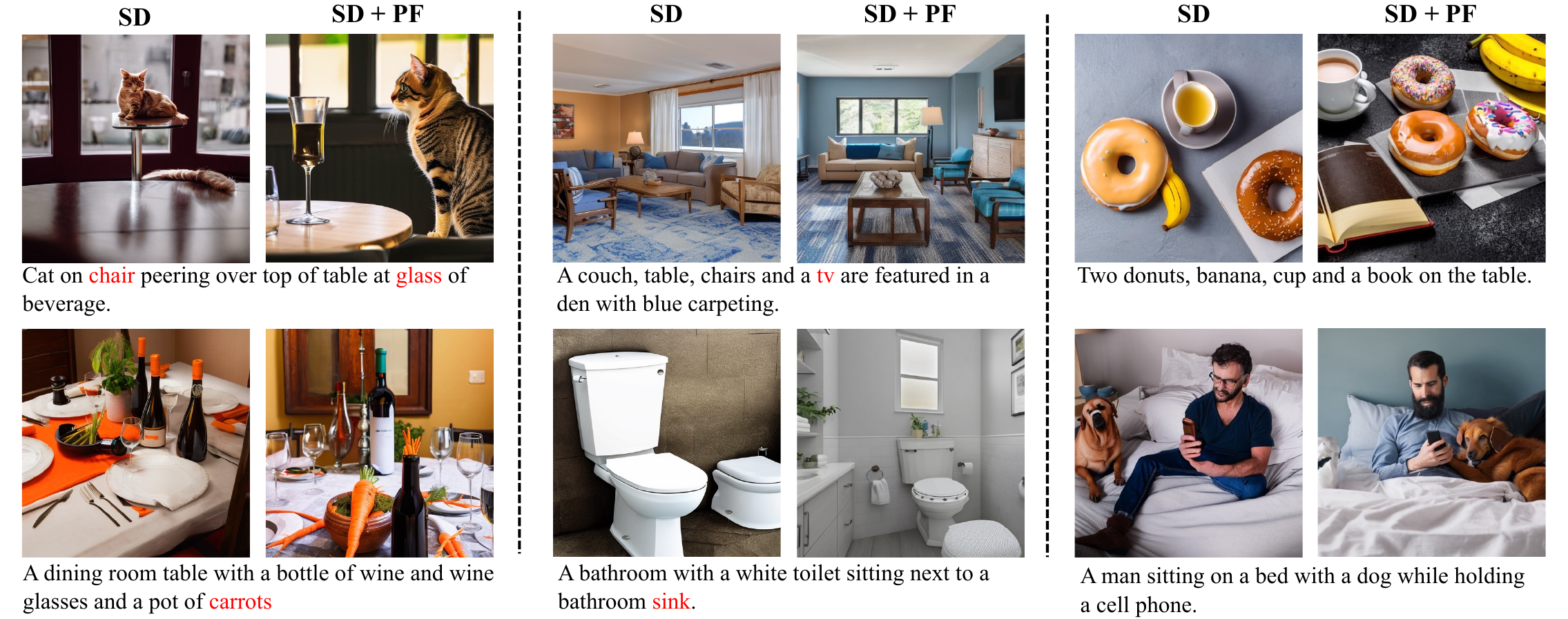}
\vspace*{-2mm}
\captionof{figure}{Sample images generated by Stable Diffusion v2.1 with and without our particle filtering algorithm (PF). The left two columns illustrate the missing object errors, and the right column illustrates the unnatural distortions. Missing objects are highlighted in \textcolor{red}{red}.}
\label{fig:teaser}
    \end{center}
}]
}

\maketitle
\input{sec/0_abstract}    
\input{sec/1_intro}
\input{sec/2_related_work}
\input{sec/3_methodology}
\input{sec/4_experiment}
\input{sec/5_conclusion}

{
    \small
    \bibliographystyle{ieeenat_fullname}
    \bibliography{main}
}

\input{sec/X_suppl}

\end{document}

%% file: preamble.tex
\usepackage[dvipsnames, table]{xcolor}


%% file: sec/0_abstract.tex
\begin{abstract}
Despite diffusion models' superior capabilities in modeling complex distributions, there are still non-trivial distributional discrepancies between generated and ground-truth images, which has resulted in several notable problems in image generation, including missing object errors in text-to-image generation and low image quality. Existing methods that attempt to address these problems mostly do not tend to address the fundamental cause behind these problems, which is the distributional discrepancies, and hence achieve sub-optimal results. In this paper, we propose a particle filtering framework that can effectively address both problems by explicitly reducing the distributional discrepancies. Specifically, our method relies on a set of external guidance, including a small set of real images and a pre-trained object detector, to gauge the distribution gap, and then design the resampling weight accordingly to correct the gap. Experiments show that our methods can effectively correct missing object errors and improve image quality in various image generation tasks. Notably, our method outperforms the existing strongest baseline by 5$\%$ in object occurrence and 1.0 in FID on \texttt{MS-COCO}. Our code is available at \url{https://github.com/UCSB-NLP-Chang/diffusion_resampling.git}.

\end{abstract}

%% file: sec/1_intro.tex
\section{Introduction}
Diffusion models have achieved impressive success in generating high-quality images in various image-generation applications \cite{sohl-dickstein15, ho2020denoising, song2021scorebased, song2021denoising, karras2022elucidating, dhariwal2021diffusion, rombach2022high, ramesh2022hierarchical, saharia2022photorealistic, Avrahami2022blended, lugmayr2022repaint, zhang2023adding, zhang2023coherent, ruiz2023dreambooth, gal2022image, hertz2022prompttoprompt, meng2022sdedit, kawar2023imagic, brooks2023instructpix2pix, wu2022uncovering, nichol2022glide, mou2023t2iadapter, song2023consistency}. Such success can be ascribed to diffusion models' superior capability in modeling complex data distributions and thus low distributional discrepancies between the real and generated images. However, the diffusion generation process still introduces distributional discrepancies in modeling data distributions, due to the limited representation power of the denoising network, the errors introduced in discretizing and numerically solving the continuous ODE or SDE denoising trajectories \cite{kim2023refining, xu2023restart}, \emph{etc}.

Such distributional discrepancies have led to several notable problems in image generation. For example, in text-to-image generation, diffusion models suffer from the \emph{missing object errors}, \emph{i.e.}, objects that are mentioned in the input text are sometimes not generated \cite{feng2023trainingfree, wu2023harnessing, marcus2022preliminary}. As shown in Figure~\ref{fig:teaser} top-left, given the input description \emph{`Cat on chair peering over top of table at glass of beverage'}, the stable diffusion model \cite{rombach2022high}, a powerful text-to-image model, sometimes fails to generate the glass of beverage.

Another well-known problem in diffusion-based image generation is the \emph{low image quality}, such as artifacts and unnatural distortions. In Figure~\ref{fig:teaser} bottom-right, given the input description \emph{`A man sitting on a bed with a dog while holding a cell phone'}, a sample image generated by stable diffusion contains a person with a missing left lower leg.

Since the fundamental cause behind these problems is the distribution gap between the generated and real images, an effective remedy to these problems should be to close the distribution gap. However, the existing research attempts to address these problems either do not aim to close the gap, or are not effective in doing so. For example, to address the missing object errors, some previous works designed cross-attention mechanism that controls how each token in the caption should be attended so that no object is ignored \citep{feng2023trainingfree, wu2023harnessing, balaji2023ediffi, lian2023llmgrounded}, but they do not aim to reduce the distributional gap, so their effects on generating complete objects are limited \citep{karthik2023dont}, and their over-emphasis on improving object occurrence could lead to quality degradation. To address the low image quality issue, \citet{kim2023refining} introduces a discriminator to rectify the score function in the diffusion generation process, which is still subject to errors due to numerically solving the ODE/SDE denoising trajectories.

In this paper, we aim to develop a sampling-based framework that can effectively address both issues above by explicitly reducing the distributional discrepancies. We first present some findings of our initial exploration, which shows that sampling-based methods turn out to be a more effective way of modifying the output distribution of generated images and achieving desired generation properties than the more complicated baselines. Based on these findings, we design a particle-filtering framework \citep{Doucet_Freitas_Gordon_2011, Chopin_Papaspiliopoulos_2020} that allows us to approximately sample from the ground-truth distribution. Specifically, our method relies on a set of external guidance, including a small set of real images and a pre-trained object detector, to gauge the distribution gap, and then design the resampling weight accordingly.

We evaluate our approach on text-to-image generation, unconditional and class-conditioned generation. On existing text prompt benchmarks \citep{wu2023harnessing, lin2014microsoft}, our approach outperforms all the competitive specialized text-to-image methods in improving faithfulness to input text and image quality, as illustrated in Figure~\ref{fig:teaser}. Notably, our method outperforms the existing strongest baseline by 5$\%$ in object occurrence and 1.0 in FID on \texttt{MS-COCO}. On \texttt{ImageNet-64}, our method achieves a state-of-the-art FID of 1.02 for class-conditioned generation, outperforming the strong baseline that uses discriminator guidance \citep{kim2023refining}.

%% file: sec/2_related_work.tex
\section{Related Works}
\label{sec:related-work}

\textbf{Faithful Text-to-image Generation.}\quad
Recent studies highlight faithfulness as a key challenge in text-to-image diffusion models \citep{feng2023trainingfree, marcus2022preliminary, wu2023harnessing, chefer2023attendandexcite, cho2023dalleval, lee2023aligning, chen2023trainingfree, ma2023directed, feng2023layoutgpt, liu2023detector},
including missing objects, mistakenly bound attributes, wrong locations, \emph{etc.}
To address this challenge, existing work modifies the generation process to separately focus on each aspect in the caption and later combine the outputs of each part \citep{liu2023compositional, du2023reduce, wang2023compositional}.
For example, \citet{wu2023harnessing} and \citet{feng2023trainingfree} modify the cross-attention between image and text to separately attend to each noun phrase in the text, and combine attention outputs. Several works leverage layout information to increase objects' attention weights in specified image regions \citep{wu2023harnessing, lian2023llmgrounded, balaji2023ediffi}. \citet{karthik2023dont} also employs the sample selection idea, but their selection is only performed at the final step, and does not aim to approach the true distribution, whereas our method reduces the distributional gap at each denoising step. Moreover, our method can extend beyond text-to-image generation.

\vspace{0.05in}
\noindent
\textbf{Particles in Diffusion Generation.}\quad
Particle filtering has been applied in diffusion generation to obtain samples from a target distribution \citep{trippe2023diffusion, wu2023practical, dou2024diffusion}. However, these works focus on re-purposing an unconditional diffusion model as a conditional model, rather than reducing the gap with the ground-truth distribution. \citet{corso2023particle} proposes particle guidance to increase the diversity among generated samples. Their method modifies the score function with a joint-particles potential, which is orthogonal to our work.

\vspace{0.05in}
\noindent
\textbf{Diffusion Generation with Discriminator.}\quad
Discriminators have been used to improve diffusion generation. \citet{xiao2022tackling} uses a GAN to approximate multimodal distributions in the denoising process to reduce the number of denoising steps. \citet{kang2023observationguided} trains diffusion models with the additional objective of fooling a discriminator to improve image quality when the number of denoising steps is small. \citet{kim2023refining} leverages a discriminator to generate samples closer to ground-truth distribution. However, their discriminator is used to modify the model score, whereas ours is used to resample images, which is less affected by the discretization errors of the sampling algorithm.

%% file: sec/3_methodology.tex
\section{Methodology}

\subsection{Background and Notation}

Throughout this section, we use upper-case letters, \e{\bm X}, to denote random vectors, and lower-case letters, \e{\bm x}, to denote deterministic vectors.
We use the colon notation, \e{\bm X_{1:T}}, to denote a set of variables ranging from \e{\bm X_1} to \e{\bm X_T}.

In this work, we will primarily focus on text-to-image diffusion models, which, given an input text description denoted as \e{\bm C}, generate images that satisfy the text description via a denoising process that generates a sequence of noisy images from \e{\bm X_T} down to \e{\bm X_0}, where \e{\bm X_0} represents the clean image, and \e{\bm X_t} represents the image corrupted with Gaussian noise. The noise is larger with a greater \e{t}, and with sufficiently large \e{T}, \e{\bm X_T} approaches pure Gaussian noise. The denoising process can be formulated as follows:
\begin{equation}
    \small
    q(\bm X_{0:T} | \bm C) = q(\bm X_T) \prod_{t=0}^{T-1} q(\bm X_{t} | \bm X_{t+1}, \bm C),
    \label{eq:denoise_markov}
\end{equation}
where \e{q(\bm X_T)} follows the pure Gaussian distribution. Different denoising algorithms would induce different transitional probabilities \e{q(\bm X_{t} | \bm X_{t+1}, \bm C)}. Since our work does not rely on specific choices of denoising algorithms, we will leave \e{q(\bm X_{t} | \bm X_{t+1}, \bm C)} abstract throughout this section.

In the following, we will use \e{p(\bm X_t | \bm C)} to denote the ground-truth distribution, \emph{i.e.,} real images corrupted with Gaussian noise, and \e{q(\bm X_t | \bm C)} as the distribution produced by the actual denoising network. Due to the learning capacity of the denoising network and approximation errors, \e{q(\bm X_t | \bm C)} can differ from \e{p(\bm X_t | \bm C)}.

\subsection{Problem Formulation}

The general goal of this paper is to reduce the gap between  \e{q(\bm X_t | \bm C)} and \e{p(\bm X_t | \bm C)}, with the help of some external guidance, in order to address the following two prominent diffusion generation errors.
\begin{itemize}
    \item \textbf{Missing Object Errors:} The generated images sometimes miss the objects mentioned in the text input. Formally, we introduce two bag-of-object variables. One is the \textit{object mention} variable, denoted as \e{\bm O_{C}}, whose \e{i}-th element, \e{O_{Ci}}, equals one if object \e{i} is mentioned in the input text \e{\bm C} and zero otherwise. The other is the \textit{object occurrence} variable, denoted as \e{\bm O_{X}}, whose \e{i}-th element, \e{O_{Xi}}, equals one if object \e{i} occurs in the image \e{\bm X_0} and zero otherwise. Then the missing object errors refer to the case where \e{O_{Ci} = 1} but \e{O_{Xi} = 0}.
    \item \textbf{Low Image Quality:} The generated images can sometimes suffer from unnatural distortion and texture. The low image quality problem can be ascribed as the general distribution gap between \e{p(\bm X_0)} and \e{q(\bm X_0)}.
\end{itemize}
We consider the following two types of external guidance to correct the generation errors.
\begin{itemize}
    \item An \textbf{object detector} which can predict the probability that a certain object appears in the image, \emph{i.e.} \e{\hat{p}(O_{Xi} = 1 | \bm X_0)}, which provides useful information in correcting the missing object errors.
    \item A small set of \textbf{real images}, \e{\mathcal{D}}, either with or without text captions, which is useful in gauging the distribution gap.
\end{itemize}
In the following, we will develop different methods under different availability of external guidance.

\subsection{Initial Exploration: A Naive Approach}
\label{sec:simple-approach}

As an initial exploration, we first focus on the subproblem that uses the object detector to correct the missing object errors. We will start with a naive sample selection approach, consisting of two steps. First, we use the diffusion model to generate \e{K} samples, \e{\bigcup_{k=1}^K\{\bm x_0^{(k)}\}}. Second, we simply select the image with the best object occurrence probability of the objects that are mentioned in the text, \emph{i.e.,}
\begin{equation}
    \small
    \max_{\bm x_0^{(k)}} \prod_{i: O_{Ci}=1} \hat{p}(O_{Xi} = 1 | \bm X_0 = \bm x_0^{(k)}).
    \label{eq:simple-selection-criterion}
\end{equation}
We dub the approach as \textsc{Objectselect}, whose implementation details are elaborated in Appendix \ref{append:t2i}.

We run \textsc{Objectselect} on two benchmark datasets (\texttt{GPT-Synthetic} \citep{wu2023harnessing} and \texttt{MS-COCO} \citep{lin2014microsoft}) with complex text descriptions, together with four baselines that also use object detectors or other object occurrence feedback to improve the object occurrence of the generated images. The object occurrence ratios of all the methods are shown in Figure \ref{fig:t2i-results}. As can be observed, \textsc{Objectselect} can significantly outperform the other baselines, some of which are much more complicated. More details of this experiment can be found in Section \ref{subsec:exp-t2i}.

Admittedly, despite the effectiveness of \textsc{Objectselect}, this algorithm comes with obvious limitations. First, it only addresses the object occurrence issue and does not aim to approach the ground-truth conditional distribution \e{p(\bm X_0 | \bm C)}, so the selected images may compromise in quality. Second, the success of this approach relies on generating a sufficiently large number of samples to be able to find a few with good object occurrence, so the sample efficiency is low.
However, this experiment does provide us with an important insight: Performing sampling on multiple generation paths of diffusion models turns out to be a direct and effective way of modifying the generation distribution of diffusion models. Similar observations are also made in previous works \citep{karthik2023dont, ramesh2021zeroshot, samuel2023generating}. If \textsc{Objectselect} can already do so well, can we design an even more effective sampling-based algorithm that overcomes its limitations and can simultaneously correct for the missing object errors and improve image quality?

\subsection{A Particle Filtering Framework}
Particle filtering has long been an effective Monte-Carlo sampling framework for approximately generating random samples from a target distribution \citep{Doucet_Freitas_Gordon_2011, Chopin_Papaspiliopoulos_2020}, and has been successfully applied to diffusion models \citep{wu2023practical, trippe2023diffusion}. Inspired by this, we will explore how to incorporate the external guidance into the particle filtering framework, to achieve a target distribution of \e{p(\bm X_0 | \bm C)}.

Our particle filtering framework obtains samples of \e{\bm X_{0:T}} in reverse order. It first generates \e{K} samples of \e{\bm X_T}, denoted as \e{\{\bm x_T^{(k)}\}}, conditional on which \e{K} samples of \e{\bm X_{T-1}} are derived. This process continues until \e{K} samples of \e{\bm X_{0}} are obtained. Specifically, samples of \e{\bm X_t}, \e{\{\bm x^{(k)}_{t}\}}, are derived from samples of \e{\bm X_{t+1}}, \e{\{\bm x_{t+1}^{(k)}\}}, via the following two steps.

\noindent
\textbf{Step 1: Proposal.} \quad For each sample \e{\bm x^{(k)}_{t+1}}, propose a sample of \e{\bm X_{t}} based on a proposal distribution \e{r(\bm X_t | \bm X_{t+1} = \bm x^{(k)}_{t+1}, \bm C)}. This sample is denoted as \e{\tilde{\bm x}^{(k)}_t}.

\noindent
\textbf{Step 2: Resampling.} \quad Given all the sample pairs \e{\bigcup_{k=1}^K \{(\bm x^{(k)}_{t+1}, \tilde{\bm x}^{(k)}_t)\}}, resample from this set \e{K} times (with replacement) with a probability proportional to a weight function, \e{w(\bm x^{(k)}_{t+1}, \tilde{\bm x}^{(k)}_t | \bm C)}, and then only keep the latter in each sample pair. The resulting samples become \e{\{\bm x_{t}^{(k)}\}}.

\begin{figure}
  \centering
  \begin{subfigure}{0.9\linewidth}
    \includegraphics[width=\linewidth]{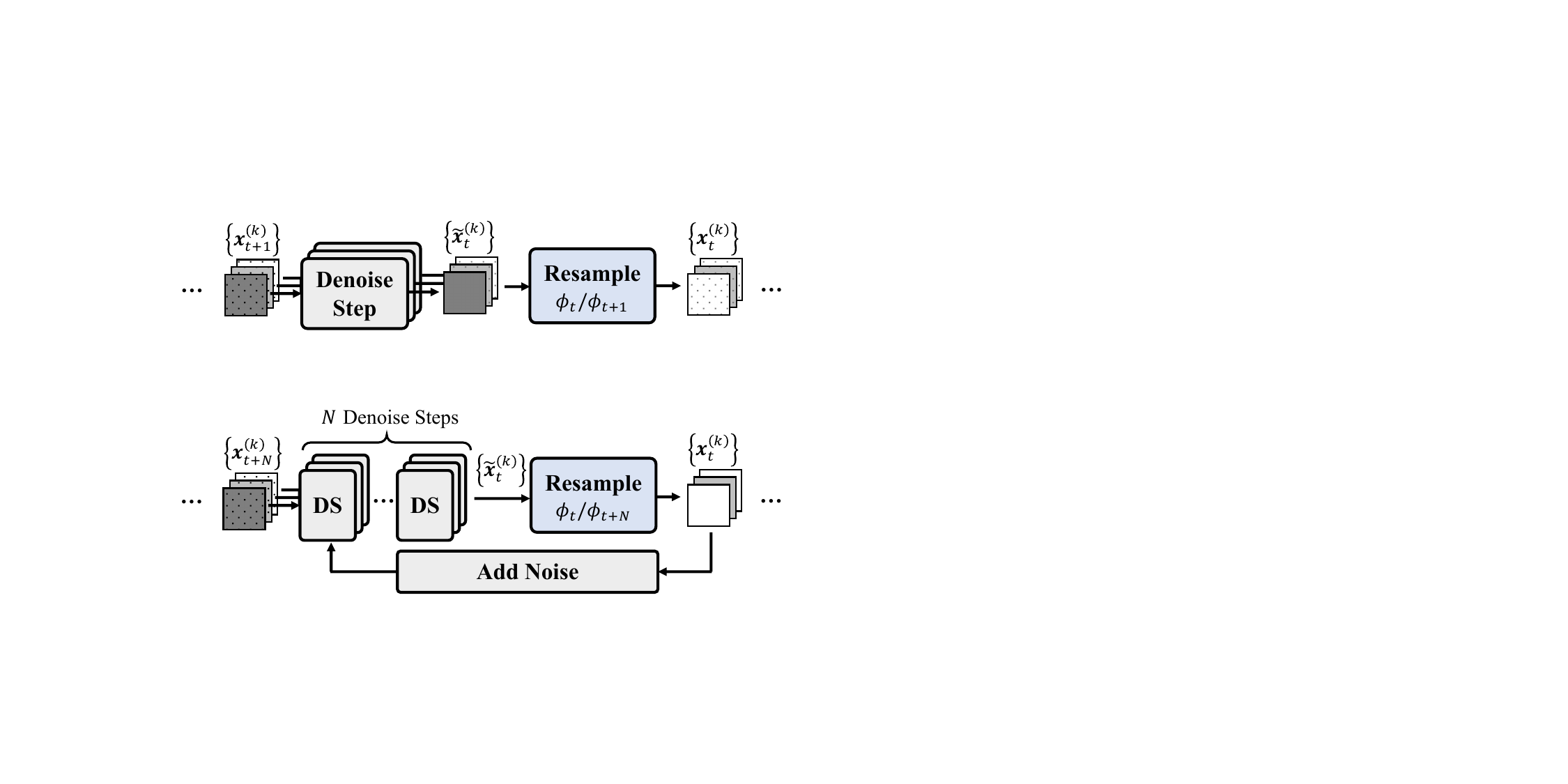}
    \caption{Generic particle filtering framework.}
    \label{fig:pf-f1}
  \end{subfigure}
  \begin{subfigure}{0.9\linewidth}
    \includegraphics[width=\linewidth]{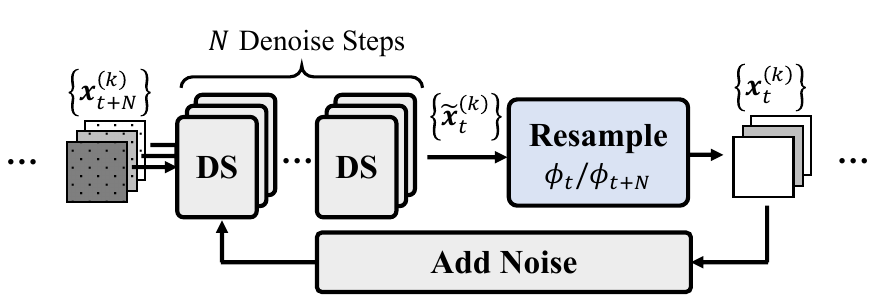}
    \caption{Particle filtering framework with the Restart sampler.}
    \label{fig:pf-f2}
    \vspace{-2mm}
  \end{subfigure}
  \caption{Illustration of our particle filtering framework.}
  \vspace{-3mm}
  \label{fig:pf}
\end{figure}

Therefore, the key to the design of a particle filtering algorithm involves designing 1) the proposal distribution \e{r(\bm X_t | \bm X_{t+1}, \bm C)}, and 2) the resampling weight \e{w(\bm X_{t+1}, \bm X_t | \bm C)}. Consistent with the design principles in \citet{wu2023practical}, we adopt the following design:
\begin{equation}
\small
\begin{aligned}
    & r(\bm X_t | \bm X_{t+1}, \bm C) = q(\bm X_t | \bm X_{t+1}, \bm C), \\
    & w(\bm X_{t+1}, \bm X_t | \bm C) = \frac{\phi_t(\bm X_t | \bm C)}{\phi_{t+1} (\bm X_{t+1} | \bm C)}. \\
\end{aligned}
\label{eq:particle_design}
\end{equation}
The first line essentially means that the proposal step becomes the denoising step of the diffusion model. Denote the distribution that \e{\{\bm x_{t}^{(k)}\}} follows as \e{v(\bm X_t | \bm C)}.
Then it can be easily shown that, if we follow the design in Eq.~\eqref{eq:particle_design},
\begin{equation}
    \small
    v(\bm X_t | \bm C) = q(\bm X_t | \bm C) \phi_t(\bm X_t | \bm C).
\end{equation}
The proof is provided in Appendix~\ref{append:proof}. In other words, \e{\phi_t(\bm X_t | \bm C)} can be interpreted as a correction term that modifies the distribution of \e{\bm X_t}. If we could set
\begin{equation}
    \small
    \phi_t(\bm X_t | \bm C) = \frac{p(\bm X_t | \bm C)}{q(\bm X_t | \bm C)},
    \label{eq:likelihood_ratio}
\end{equation}
then we can ideally achieve \e{v(\bm X_t | \bm C) = p(\bm X_t | \bm C), \forall t}, which means the final generated image will follow the real image distributions. Figure \ref{fig:pf-f1} illustrates the overall process.

Now the question is how to compute the conditional likelihood ratio in Eq.~\eqref{eq:likelihood_ratio}. Sections~\ref{subsec:method1} and \ref{subsec:method2} introduce two methods with different requirements on external guidance.

\subsection{A Discriminator-Based Approach}
\label{subsec:method1}

\begin{figure}
\centering
  \begin{subfigure}{0.85\linewidth}
    \includegraphics[width=\linewidth]{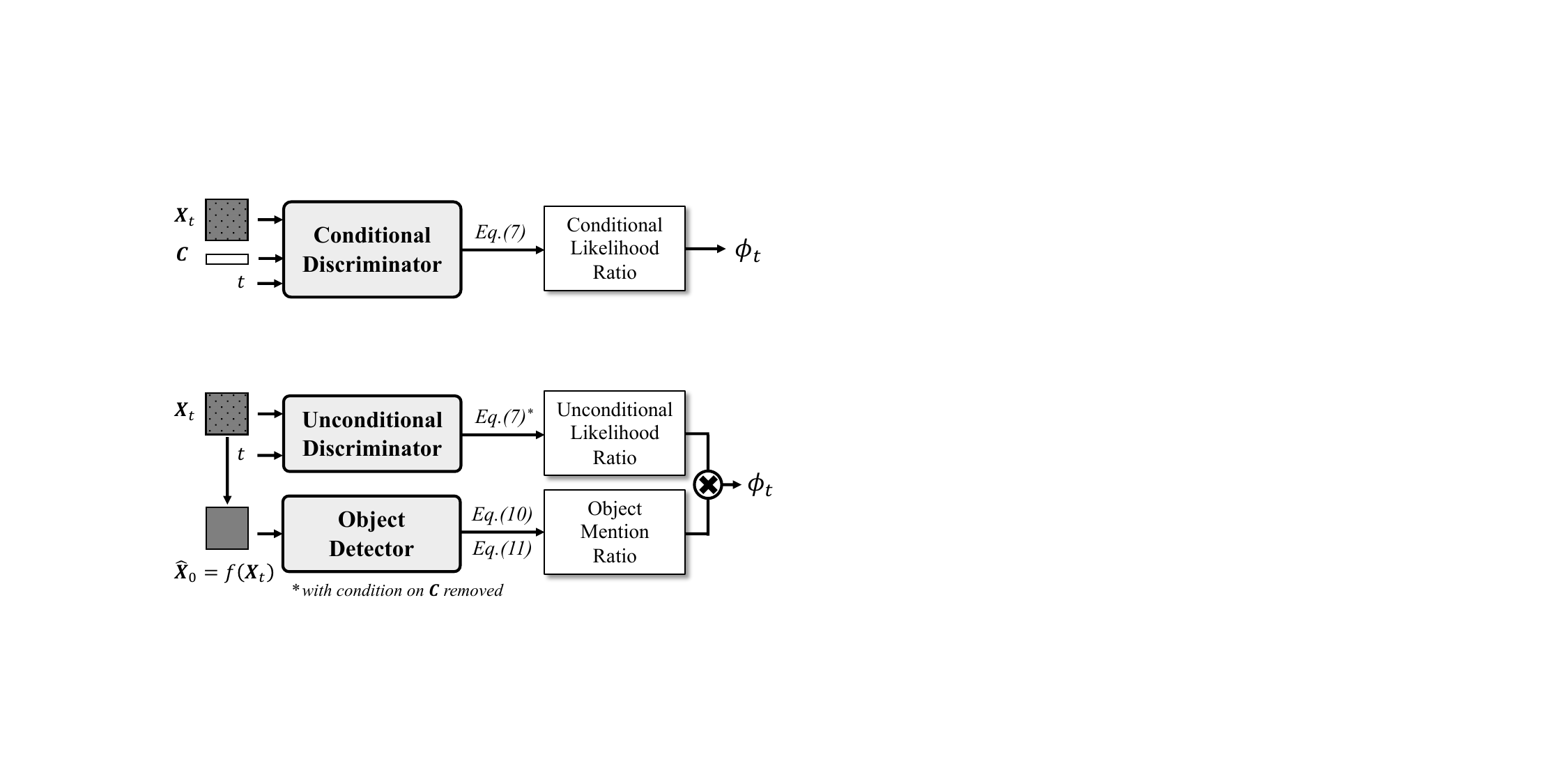}
    \caption{Discriminator-based approach.}
    \vspace{1mm}
    \label{fig:weight-f1}
  \end{subfigure}
  \begin{subfigure}{0.85\linewidth}
    \includegraphics[width=\linewidth]{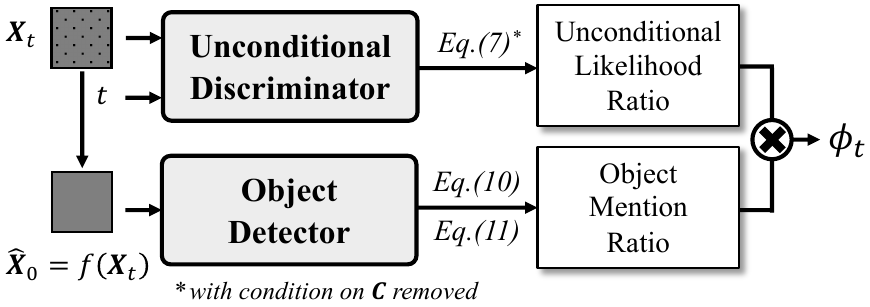}
    \caption{Hybrid approach.}
    \label{fig:weight-f2}
    \vspace{-2mm}
  \end{subfigure}
  \caption{Calculation of the correction term \e{\phi_t(\bm X_t | \bm C)}.}
  \label{fig:weight}
  \vspace{-3mm}
\end{figure}

Our first method, shown in Figure \ref{fig:weight-f1}, only utilizes a small set of real images with captions as the external guidance. It involves training a discriminator that discriminates between real and generated examples. Specifically, given a text description \e{\bm C}, real samples of \e{\bm X_t} are obtained by corrupting real images that correspond to \e{\bm C} with Gaussian noise; fake samples of \e{\bm X_t} are obtained by first generating images using the diffusion model with \e{\bm C} as the text condition, and then corrupting the generated images with Gaussian noise.

Given the real and fake samples, we train a \emph{conditional} discriminator \e{d(\bm X_t| \bm C; t)} by minimizing the canonical discrimination loss 
\begin{equation}
\small
\begin{aligned}
\mathcal{L} =  \mathbb{E}_{\bm C, t}\big[&\mathbb{E}_{\bm X_t \sim p(\bm X_t | \bm C)}[-\log d(\bm X_t| \bm C; t)] \\
+ & \mathbb{E}_{\bm X_t \sim q(\bm X_t | \bm C)}[-\log (1-d(\bm X_t| \bm C; t))]\big].
\end{aligned}
\label{eq:loss}
\end{equation}
It has been shown \cite{goodfellow2014generative} that the minimizer of \eqref{eq:loss}, denoted as \e{d^*(\bm X_t| \bm C; t)}, can be used to compute the conditional likelihood ratio as:
\begin{equation}
\small
    \frac{p(\bm X_t | \bm C)}{q(\bm X_t | \bm C)} \approx \frac{d^*(\bm X_t | \bm C; t)}{1 - d^*(\bm X_t | \bm C; t)}.
    \label{eq:discriminator}
\end{equation}
Theoretically, the discriminator-based approach can correct any distributional discrepancies between real and generated images. However, due to the limitation in representation power and optimization schemes, the discriminator may fail to capture certain errors, such as the missing object errors. Next, we will introduce an alternative approach with a more fine-grained correction of different error types.

\subsection{A Hybrid Approach}
\label{subsec:method2}

The hybrid approach (in Figure \ref{fig:weight-f2}) uses both the object detector and real image set as the external guidance to address the missing object errors and poor image quality respectively. Formally, according to the Bayes rule, we decompose the conditional likelihood ratio in Eq.~\eqref{eq:likelihood_ratio} as follows:
\begin{equation}
\small
\begin{aligned}
    \frac{p(\bm X_t | \bm C)}{q(\bm X_t | \bm C)} &\stackrel{\text{\ding{172}}}{=} \frac{p(\bm X_t | \bm C, \bm O_C)}{q(\bm X_t | \bm C, \bm O_C)}\\
    &= \frac{q(\bm C, \bm O_C)}{p(\bm C, \bm O_C)} \cdot \frac{p( \bm X_t)}{q ( \bm X_t)} \cdot \frac{p(\bm O_C | \bm X_t)}{q (\bm O_C | \bm X_t)} \cdot \frac{p(\bm C | \bm O_C, \bm X_t)}{q (\bm C | \bm O_C, \bm X_t)}.
\end{aligned}
\label{eq:weight_decomp}
\end{equation}
Equality \ding{172} is due to the fact that the information in \e{\bm O_C}, by definition, comes entirely from \e{\bm C}. The right-hand side of Eq.~\eqref{eq:weight_decomp} consists of four terms. The first term is about the prior distribution of text conditions, which should equal one because the text prior is not impacted by the denoising process. The second term addresses the general distributional discrepancies in images, which accounts for the \emph{low image quality} issue. The third term accounts for the missing object errors. The fourth term accounts for any other inconsistencies between the image and text, such as inconsistent object characteristics, mispositioned objects, \emph{etc.} Since our current focus is on correcting missing object errors and improving image quality, we will focus on estimating the second and third terms, which we will refer to as the \textit{unconditional likelihood ratio} and \textit{object mention ratio} respectively.

\vspace{0.05in}
\noindent
\textbf{Estimating the unconditional likelihood ratio.} \quad
The unconditional likelihood ratio can be estimated the same way as in Section~\ref{subsec:method1}, except that the discriminator should be replaced with an unconditional one without \e{\bm C}, hence denoted as \e{d(\bm X_t; t)}. Accordingly, we can replace the discriminator in  Eq.~\eqref{eq:discriminator} with \e{d^*(\bm X_t; t)} to compute the unconditional likelihood ratio. As a result, this hybrid approach does not require that the image set comes with captions.

\vspace{0.05in}
\noindent
\textbf{Estimating the object mention ratio.} \quad
Assuming conditional independence of different dimensions of \e{\bm O_C}, the object mention ratio can be further factorized as follows:
\begin{equation}
\small
\begin{aligned}
    &\frac{p(\bm O_C | \bm X_t)}{q (\bm O_C | \bm X_t)} = \frac{\prod_i p(O_{Ci} | \bm X_t)}{\prod_i q(O_{Ci} | \bm X_t)}\\
    &= \frac{\prod_{i: O_{Ci}=1} p(O_{Ci}=1 | \bm X_t)}{\prod_{i: O_{Ci}=1} q(O_{Ci} = 1 | \bm X_t)} \cdot \frac{\prod_{i: O_{Ci}=0} p(O_{Ci}=0 | \bm X_t)}{\prod_{i: O_{Ci}=0} q(O_{Ci} = 0 | \bm X_t)}.
\end{aligned}
\label{eq:object_occur_decomp}
\end{equation}
Essentially, Eq.~\eqref{eq:object_occur_decomp} divides the objects into two groups, ones that are mentioned in the caption, and ones that are not. So the first term corrects for the \emph{missing object errors} and the second term corrects for the \emph{false generation} (\emph{i.e.}, generating objects that are not mentioned in the caption). Since the main concern of diffusion generation is the former, we will focus on computing the first term.

The numerator, \e{p(O_{Ci} = 1 | \bm X_t)}, measures, pretending that the noisy image \e{\bm X_t} were corrupted from a real image and that the real image had a caption, what is the probability that the caption would mention object \e{i}. Since the caption almost always reflects what is in the image, this probability can be easily approximated by running the object detector on the predicted clean image from \e{\bm X_t}. Formally, denote \e{f(\bm X_t)} as the one-step prediction of the clean image by the diffusion model, then
\begin{equation}
    \small
    p(O_{Ci} = 1 | \bm X_t) \approx \hat{p}(O_{Xi} = 1 | f(\bm X_t)).
    \label{eq:numerator}
\end{equation}
A more rigorous derivation is provided in Appendix~\ref{append:resample-weight}.

The denominator, \e{q(O_{Ci} = 1 | \bm X_t)}, measures, pretending that the noisy image \e{\bm X_t} were generated by the (imperfect) diffusion model conditional on an input description \e{\bm C}, what is the probability that the input text had mentioned object \e{i}. Unlike the numerator, if an object \e{i} does not appear in the image, there is still a non-zero chance that the caption had mentioned the object \e{i}, considering the diffusion model may miss the object. As formally derived in Appendix~\ref{append:resample-weight}, this probability can be approximated as
\begin{equation}
    \small
    \begin{aligned}
        q(O_{Ci} = 1 &| \bm X_t)  \approx \hat{p}(O_{Xi} = 1 | f(\bm X_t)) \\
        & + \hat{p}(O_{Xi} = 0 | f(\bm X_t)) \frac{(1-\kappa_{it})\pi_{it}}{(1-\kappa_{it})\pi_{it} + 1 - \pi_{it}},
    \end{aligned}
    \label{eq:denominator}
\end{equation}
where \e{\pi_{it}} is a hyper-parameter between \e{0} and \e{1}; and
\begin{equation}
\small
    \kappa_{it} = \frac{\mathbb{E}_{\bm X_t, \bm C \sim q(\bm X_t, \bm C)}[\mathbbm{1}(\hat{O}_{Xit} = 1 \wedge O_{Ci}=1)]}{\mathbb{E}_{\bm C \sim q(\bm C)}[\mathbbm{1}(O_{Ci}=1)]}
    \label{eq:kappa}
\end{equation}
denotes the percentage of occurrence of object \e{i} in the image predicted from \e{\bm X_t} when the text description mentions object \e{i}. \e{\hat{O}_{Xit}} denotes the (estimated) occurrence of object \e{i} in the clean image predicted from \e{\bm X_t}, \emph{i.e.} \e{f(\bm X_t)}, which is computed by passing \e{f(\bm X_t)} to the object detector and then see whether the output exceeds the threshold \e{0.5}. \e{\mathbbm{1}(\cdot)} denotes the indicator function.

To compute \e{\kappa_{it}}, we would need to run an initial generation round, where we feed a set of text descriptions to the diffusion model (\emph{without} particle filtering) and generate a set of images. The numerator of Eq.~\eqref{eq:kappa} is computed by counting how many text-image pairs mention object \e{i} in the text and the corresponding \e{\hat{O}_{Xit}} equals one; the denominator is computed by counting how many text descriptions mention object \e{i}. Although the initial generation round introduces extra computation, it only needs to perform once.

\vspace{0.05in}
\noindent
\textbf{Summary.} \quad Our hybrid approach computes \e{\phi_t(\bm X_t | \bm C)} as
\begin{equation}
    \small
    \phi_t(\bm X_t | \bm C) = \frac{p( \bm X_t)}{q ( \bm X_t)} \cdot \frac{\prod_{i: O_{Ci}=1} p(O_{Ci}=1 | \bm X_t)}{\prod_{i: O_{Ci}=1} q(O_{Ci} = 1 | \bm X_t)},
    \label{eq:hybrid-summary}
\end{equation}
where the first term is computed by training an unconditional discriminator (similar to Eq.~\eqref{eq:discriminator}), and the second term is computed from Eqs.~\eqref{eq:numerator} and \eqref{eq:denominator} using the object detector. Algs.~\ref{alg:pf} and \ref{alg:calculate_phi} in Appendix describe the particle filtering algorithm and the calculation of \e{\phi_t(\bm X_t | \bm C)} respectively.

\subsection{Generalization to Other Generation Settings}
\label{subsec:remarks}

The proposed algorithm can be applied beyond text-to-image generation to generic conditional and unconditional generations by setting \e{\bm C} to other conditions or an empty set respectively. In this case, the discriminator-based approach will be used as the hybrid approach is no longer applicable.

%% file: sec/4_experiment.tex
\section{Experiments}
In this section, we will first demonstrate the effectiveness of our method on reducing the missing object errors and improving the image quality in text-to-image generation. We then evaluate our method on standard benchmarks of unconditional and class-conditioned generation. Finally, we will investigate various design choices in our framework. 

\subsection{Text-to-Image Generation}
\label{subsec:exp-t2i}

\textbf{Datasets.}\quad
We use two datasets: (1) \texttt{GPT-}\texttt{Synthetic} was introduced in \citet{wu2023harnessing} to evaluate text-to-image models' ability in generating correct objects and their associated colors and positions. It contains 500 captions, where each caption contains 2 to 5 objects in \texttt{MS-COCO} \citep{lin2014microsoft}, and objects are associated with random colors and spatial relations with other objects, \emph{e.g.,} \emph{`The tie was placed to the right of the red backpack.'}
(2) We also evaluate on the validation set of \texttt{MS-COCO}. However, we focus on the subset of complex descriptions that contain at least four objects, which results in 261 captions (details in Appendix \ref{append:implement-details}).

\vspace{0.05in}
\noindent
\textbf{Baselines.}\quad
We compare with seven baselines: \ding{182} \textsc{SD}, which is the Stable Diffusion model \citep{rombach2022high} as is;
\ding{183} \textsc{D-guidance} \citep{kim2023refining} that modifies the score function by adding an additional term \e{\nabla \log\frac{p(\bm X_t | \bm C)}{q(\bm X_t | \bm C)}}, where we use the hybrid approach in Section \ref{subsec:method2} to estimate \e{\phi_t(\bm X_t | \bm C) = \frac{p(\bm X_t | \bm C)}{q(\bm X_t | \bm C)}};
\ding{184} \textsc{Spatial-temporal} \citep{wu2023harnessing} and \ding{185} \textsc{Attend-excite} \citep{chefer2023attendandexcite} that modify each denoising step to ensure each object in the caption is attended (details in Section \ref{sec:related-work});
\ding{186} \textsc{Objectselect} in Section \ref{sec:simple-approach};
\ding{187} \textsc{TIFAselect} and \ding{188} \textsc{Rewardselect}, both in \citep{karthik2023dont}, which are similar to \ding{186} but use TIFA score \citep{hu2023tifa} and ImageReward \citep{xu2023imagereward} as the selection criteria respectively. For methods that leverage an object detector, we use DETR \citep{carion2020endtoend} with ResNet-50 backbone \citep{he2015deep}. For all sample selection methods (\ding{186}-\ding{188} and ours), we select the best image based on each method's sampling criterion for evaluation. We use SD v2.1-base for our method and baselines (only \ding{184} does not support it, for which we use v1.5). Additional results using SD v1.5 are in Appendix \ref{append:t2i-sd15-results}.

\vspace{0.05in}
\noindent
\textbf{Metrics.}\quad
We evaluate generated images using both objective and subjective metrics. For objective evaluation, we use \textbf{Object Occurrence} to measure the percentage of objects that occur in generated images over all objects mentioned in the caption. We calculate the percentage for each caption and take the average over all captions in the dataset. We use an object detector (DETR with ResNet-101 backbone) different from the one used during generation. In addition, we also calculate Fréchet inception distance (\textbf{FID}) \citep{heusel-gans} as a measure for image quality. Similar to \citet{xu2023restart}, we calculate FID on 5,000 captions in the \texttt{MS-COCO} validation set that contain at least one object.
For subjective evaluation, we recruit annotators on Amazon Mechanical Turk to compare images generated by our method and baselines. Annotators are asked to evaluate two aspects: (1) (\textbf{Object}) Given an image, annotators need to identify all objects in it. (2) (\textbf{Quality}) Given two images for the same caption, annotators need to select the one that looks more real and natural. We randomly sample 100 captions\footnote{63 $\%$ of captions in \texttt{MS-COCO} dataset contain ``person'' as an object. To evaluate on a more uniform distribution of objects, we randomly sample 25 captions that contain person and 75 captions that do not.} from each dataset for evaluation, and each image is rated by two annotators.

\vspace{0.05in}
\noindent
\textbf{Diffusion samplers.}\quad
All the methods are evaluated on the Restart sampler \citep{xu2023restart}. The only exceptions are \textsc{Spatial-temporal} and \textsc{Attend-exicte}, which use the same samplers as in their papers. The Restart sampler iterates between three steps: \ding{182} \emph{Denoise} from \e{t+N} to \e{t} using the ODE trajectory; \ding{183} \emph{Restart} by adding the noise back to the level \e{t+N}. \ding{184} \emph{Denoise} from \e{t+N} to \e{t} again, and either go back to step \ding{183} or proceed to the next iteration which denoises to \e{t-N'}. As shown in Figure \ref{fig:pf-f2}, our methods are combined with the Restart sampler by inserting the resampling module right before adding noise. We also experimented with the SDE sampler in EDM \citep{karras2022elucidating}, but we found Restart sampler generally dominates EDM on both object occurrence and image quality. We thus only focus on Restart sampler in the main paper. EDM results are shown in Appendix \ref{append:t2i-edm-results}.

\vspace{0.05in}
\noindent
\textbf{Sampling configurations.}\quad
To have a fair comparison in terms of the computation cost, we evaluate all sample selection methods (including ours) under a comparable number of function evaluations (NFE). Specifically, we generate each image with a fixed NFE and report performance when \e{K=5, 10, 15} images are generated for each caption. This ensures a fair comparison for methods with the same value of \e{K}. For non-selection methods (\textsc{SD}, \textsc{D-guidance}, \textsc{Spatial-temporal}, and \textsc{Attend-excite}), we use the original sampling configuration in their papers and thus only report a single performance on each dataset. Appendix \ref{append:t2i-sampling-config} details the computation cost for all methods.

\begin{figure}
    \centering
    \includegraphics[width=\linewidth]{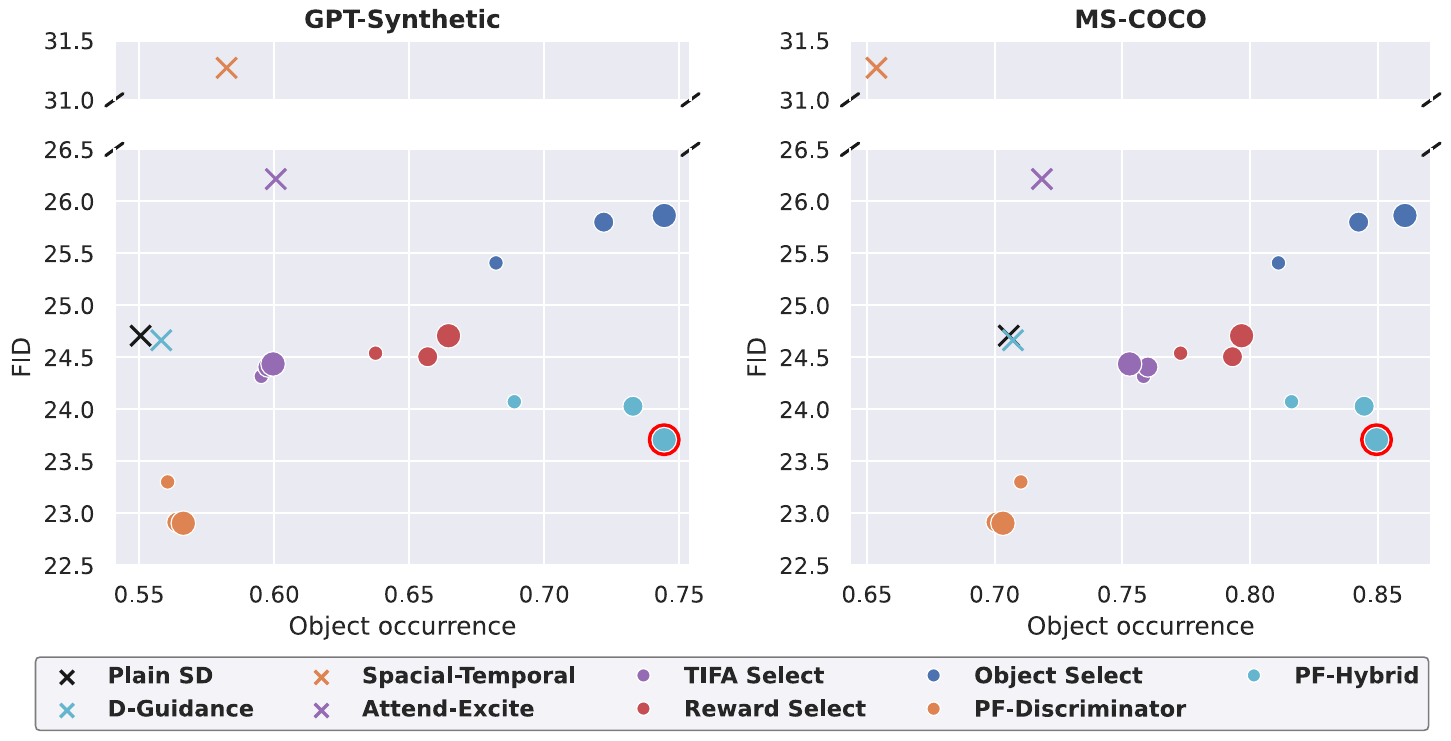}
    \vspace{-6mm}
    \caption{FID ($\downarrow$) \emph{vs.} Object occurrence ($\uparrow$) for all methods. Ideal points should scatter at the bottom right corner. Object occurrence is measured on \texttt{GPT-Synthetic} (left) and \texttt{MS-COCO} (right), and FID is measured on \texttt{MS-COCO}. \e{K=5, 10, 15} images are generated for sample selection methods, and the sizes of points indicate the value of \e{K} (larger \e{K} has larger points). The method that achieves the best combined performance is highlighted in \textcolor{red}{red}.}
    \label{fig:t2i-results}
    \vspace{-4mm}
\end{figure}

\begin{table}
\centering
\resizebox{\linewidth}{!}{
\begin{tabular}{lcccc}
\topline
\multicolumn{1}{c}{\multirow{2}{*}{\textbf{}}} & \multicolumn{2}{c}{\textbf{\texttt{GPT-Synthetic}}} & \multicolumn{2}{c}{\textbf{\texttt{MS-COCO}}} \\
\cmidrule(lr){2-3}
\cmidrule(lr){4-5}
\multicolumn{1}{c}{\multirow{2}{*}{\textbf{}}} &
Object & Quality \emph{vs.} & Object  & Quality \emph{vs.} \\
& Occur ($\uparrow$) & \textsc{PF-hybrid} ($\uparrow$) & Occur ($\uparrow$) & \textsc{PF-hybrid} ($\uparrow$) \\
\midtopline

\textsc{SD} \citep{rombach2022high} & 64.48 & -6.00 & 59.87 & -3.00 \\
\textsc{Rewardselect} \citep{karthik2023dont} & 71.97 & -3.00 & 63.25 & -3.00 \\
\textsc{Objectselect} & 70.96 & -11.00 & 67.35 & -9.00 \\
\textsc{PF-discriminator} & 68.87 & -8.00 & 59.98 & -2.00 \\
\rowcolor{gray!20}
\textsc{PF-hybrid} & 75.79 & -- & 68.13 & -- \\
\bottomline
\end{tabular}
}
\vspace{-2mm}
\caption{Human evaluation on object occurrence and image quality. `Quality' is the win rate against \textsc{PF-hybrid} (minus 50). Negative values indicate the method is worse than \textsc{PF-hybrid}.}
\vspace{-3mm}
\label{tab:t2i-subjective}
\end{table}

\vspace{0.05in}
\noindent
\textbf{Results.}\quad
Figure \ref{fig:t2i-results} shows object occurrence and FID for all methods. The performance of sample selection methods is reported for three different values of \e{K}, which are indicated by the sizes of the points in the figure. A competitive algorithm should achieve a high object occurrence rate (right) and low FID (bottom), so the more the algorithm lies to the \emph{bottom-right corner}, the more competitive the algorithm is.

There are two observations from Figure \ref{fig:t2i-results}. First, the sampling-based methods (represented by circles) generally significantly outperform the non-sampling-based ones (represented by crosses). In particular, although \textsc{Spacial-temporal} and \textsc{Attend-excite} improve object occurrence, the improvements are not as large as other sampling-based methods, and their computational costs (as shown in Appendix \ref{append:t2i-sampling-config}) are also high or on par with the sampling-based methods. Second, our \textsc{PF-hybrid} is the only algorithm that simultaneously achieves high object occurrence and low FID. \textsc{Object-select} achieves a high object occurrence, but the worst FID scores. The other proposed method, \textsc{PF-Discriminator}, achieves a very low FID, but the object occurrence is significantly compromised, which verifies our claim that conditional discriminators tend to overlook object occurrence and focus only on image quality (Section~\ref{subsec:method1}). Notably, all three methods proposed in this paper, \textsc{Object-select}, \textsc{PF-discriminator} and \textsc{-hybrid} lie at the frontier of the performance trade-off, significantly outperforming the existing baselines. Appendix \ref{append:object-category} further shows the object occurrence for each object category, which indicates that our method is particularly beneficial on small objects with fine details.

\begin{figure*}
    \centering
    \includegraphics[width=0.91\textwidth]{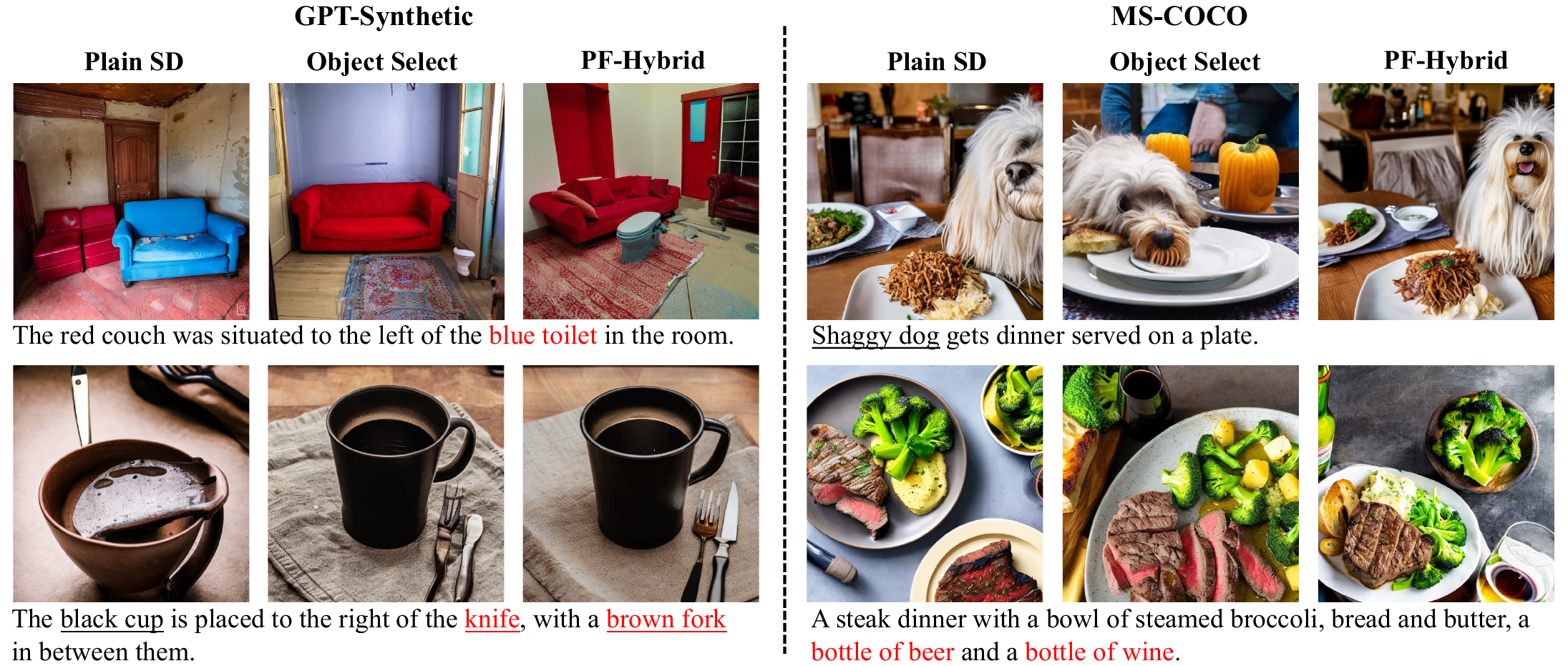}
    \vspace{-2mm}
    \caption{Visualization of generated samples. Missing objects are highlighted in \textcolor{red}{red}. Unnatural objects are highlighted with \underline{underline}.}
    \label{fig:t2i-example}
    \vspace{-3mm}
\end{figure*}

Table \ref{tab:t2i-subjective} shows the results for subjective evaluations, where object occurrence is computed the same way as the objective object occurrence. The quality score is computed as follows. Since \textsc{PF-hybrid} is the most competitive algorithm, we perform a pairwise comparison between \textsc{PF-hybrid} and each baseline. The baseline gets one point everytime a subject prefers the baseline over \textsc{PF-hybrid}. Each comparison consists of 100 pairs so a score of 50 indicates a tie. We subtract all the scores by 50 so a negative score in each baseline indicates that \textsc{PF-hybrid} is better. As can be observed, \textsc{PF-hybrid} outperforms all the baselines in terms of object occurrence and subjective quality.

Figure \ref{fig:t2i-example} visualizes some generated images by our method and baselines. As can be observed, the plain \textsc{SD} can generate natural images but tends to miss objects mentioned in the text. \textsc{Objectselect} reduces the missing object errors but could lead to unnatural objects in the image (\emph{e.g.,} knife and fork). \textsc{PF-hybrid} generates more complete objects and also improves image quality. Appendix \ref{append:image-sample} presents more examples including failure cases of our method.

\subsection{Unconditional \& Class-conditioned Generation}
\label{subsec:exp-unconditional}

\textbf{Experiment setup.}\quad
In addition to text-to-image, we evaluate two other image generation benchmarks, \texttt{FFHQ} \citep{karras2019stylebased} for unconditional generation and \texttt{ImageNet-64} \citep{deng-imagenet} for class-conditioned generation. Here, object occurrence is not applicable so we will focus on image quality as measured by FID.
We will only implement \textsc{PF-discriminator} because \textsc{PF-hybrid} is not applicable (Sectoin~\ref{subsec:remarks}).
We use the pre-trained diffusion models in \citet{karras2022elucidating} and follow \citet{kim2023refining} to train the discriminator (details in Appendix \ref{append:benchmark-discriminator-training}). For evaluation, we calculate FID on 50,000 generated images. Additional results using ADM \citep{dhariwal2021diffusion} and VP \citep{song2021scorebased} diffusion models are in Appendix \ref{append:benchmark-admvp-results}.

\vspace{0.05in}
\noindent
\textbf{Baselines and method variants.}\quad
We consider three baselines: \ding{182} The original Restart sampler; \ding{183} \textsc{D-guidance} using the discriminator to compute the correct term (following \citep{kim2023refining});
\ding{184} \textsc{D-select}, which is similar to \textsc{Objectselect} but uses the likelihood ratio \e{\frac{p(\bm X_0 | \bm C)}{q(\bm X_0 | \bm C)}} as the selection criterion (effectively an importance sampling approach to restore the ground-truth distribution). 

\vspace{0.05in}
\noindent
\textbf{Sampling configurations.}\quad
Similar to Section \ref{subsec:exp-t2i}, we evaluate \textsc{PF} and \textsc{D-select} for different values of \e{K} with a fixed NFE per image. For a fair comparison, we increase denoising steps for the original sampler and \textsc{D-guidance} to match the total NFE of the sampling-based methods.

\begin{figure}
    \centering
    \includegraphics[width=\linewidth]{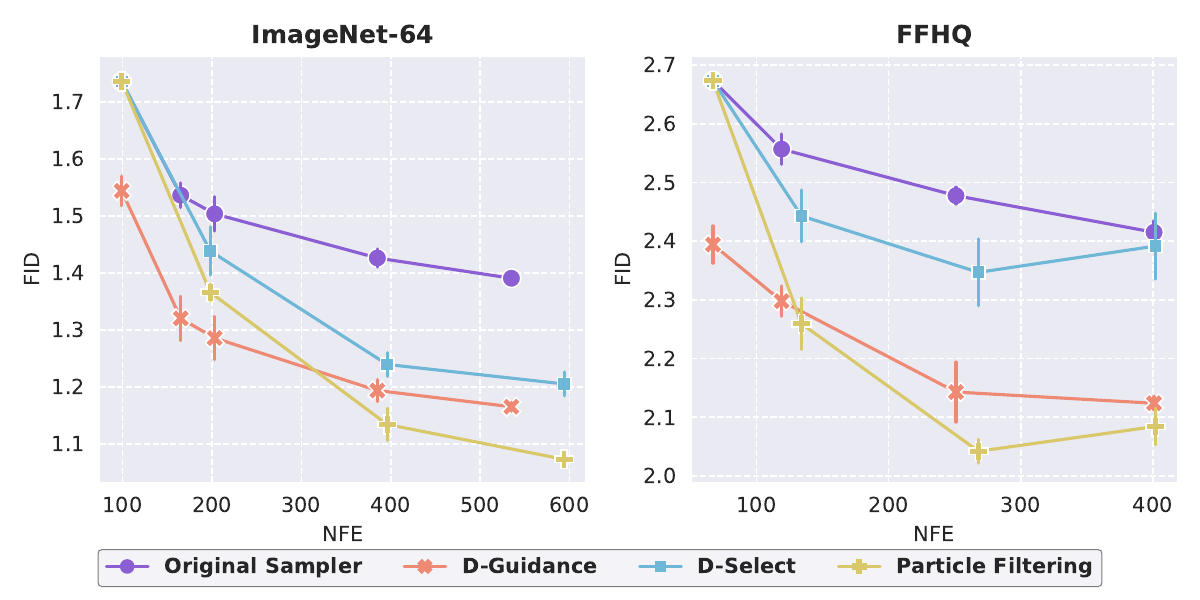}
    \vspace*{-6mm}
    \caption{FID (average of 3 runs) on \texttt{ImageNet-64} (left) and \texttt{FFHQ} (right). Error bars indicate standard deviations.}
    \label{fig:benchmark-fid}
    \vspace*{-2mm}
\end{figure}

\vspace{0.05in}
\noindent
\textbf{Results.}\quad
Figure \ref{fig:benchmark-fid} shows FID as a function of overall NFE. For \textsc{PF} and \textsc{D-select}, \e{K=2, 4, 6} images are generated.

There are two observations. First, all the methods generally improve as NFE increases, showing the effectiveness of increasing samples (for sampling-based methods) or increasing denoising steps (for non-sampling-based methods).  Second, \textsc{D-Guidance} generally performs competitively, especially at small NFEs, but our method consistently achieves the lowest FID across two datasets at large NFEs. Moreover, it is worth to note that NFE is not an adequate measure of computation cost in our setting. NFE only measures the number of evaluations of the denoising U-Net, ignoring the other compute costs such as the forward and backward passes of the discriminator. Since \textsc{D-Guidance} evaluates the discriminator at every denoising step, whereas ours only at a subset of steps with no backward passes, our method incurs 0.66$\times$ compute cost per NFE compared to \textsc{D-Guidance}. As shown in Figure \ref{fig:restart_fid_effectivenfe} in Appendix \ref{append:benchmark-effectivenfe}, if this factor is considered, our method outperforms \textsc{D-Guidance} across all compute costs.

\subsection{Ablation Study}
\label{sec:ablation}

We now investigate two important aspects of our method in text-to-image generation setting: the particle filtering algorithm and the resampling weight calculation. For each ablated version, we will evaluate on the object occurrence on \texttt{GPT-Synthetic} and \texttt{MS-COCO}, and FID on the latter.

\begin{table}
\centering
\resizebox{0.8\linewidth}{!}{
\begin{tabular}{lccc}
\topline
& \multicolumn{2}{c}{\textbf{Object Occurrence} ($\%$) $\uparrow$} & \multirow{2}{*}{\textbf{FID} $\downarrow$} \\
\cmidrule(lr){2-3}
& \texttt{GPT-Syn} & \texttt{MS-COCO} &  \\
\midrule
\rowcolor{gray!20}
\textsc{PF-hybrid} &  72.96 &  83.84 & 24.03 \\
\quad - PF&  67.16&  80.49& 24.18\\
\quad - Discriminator & 75.67 & 85.79 & 25.77 \\
\midrule
\rowcolor{gray!20}
\textsc{PF-discriminator} &  56.86 &  70.96 & 22.91 \\
\quad - PF& 57.09& 72.13&23.28\\
\quad Object only &  63.15&  77.81& 25.31\\
\bottomline
\end{tabular}
}
\vspace{-1mm}
\caption{Ablation study on the effects of particle filtering algorithm and particle weights design.}
\label{tab:ablation-t2i}
\vspace{-4mm}
\end{table}

First, we investigate the effects of \textbf{removing particle filtering}. To do that, we compare \textsc{PF-hybrid} and \textsc{PF-discriminator} with two approaches that do not involve particle filtering but only generate \e{K} images via the regular Restart sampler and select the best one with the largest value of \e{\phi_0(\bm X_0 | \bm C)}, calculated the same way as \textsc{PF-hybrid} and \textsc{PF-discriminator} respectively. Table \ref{tab:ablation-t2i} shows the performance of these two methods (rows `$-$PF'). As can be observed, removing PF hurts the performance of \textsc{PF-hybrid} on all metrics, demonstrating the importance of the particle filtering process. Removing PF hurts FID for \textsc{PF-discriminator} but does not affect object occurrence, which indicates the conditional discriminator focuses on image quality instead of object occurrence.

Second, we investigate the design choices of the \textbf{resampling weight}. We explore two variants for estimating \e{\phi_t(\bm X_t | \bm C)}. For \textsc{PF-hybrid}, we remove its unconditional discriminator and only include the object mention ratio (Eq. \eqref{eq:hybrid-summary}). For \textsc{PF-discriminator}, since we have observed that the discriminator tends to ignore missing objects, we introduce a variant to force its attention on missing objects by training the discriminator on the object detector's output probability as the feature (instead of on the images). The results in Table \ref{tab:ablation-t2i} (rows `$-$Discriminator' and `Object only' respectively) show that these two variants manage to improve object occurrence, but at the cost of higher FIDs, which highlights the needs of balancing the two objectives.

We further study other aspects in class-conditioned generation, including when the resampling is performed (\emph{i.e.,} before or after adding noise), the amount of noise being added, and the number of denoising steps per image. Notably, our \textsc{PF} method with Restart sampler achives the state-of-the-art FID of 1.02 on \texttt{ImageNet-64} when 4 images are generated with 165 NFE (details in Appendix \ref{append:ablation}).

%% file: sec/5_conclusion.tex
\section{Conclusion}
In this paper, we propose a sampling-based approach using particle filtering to correct diffusion generation errors and reduce discrepancies between model-generated and real data distributions. Experiments on text-to-image, unconditional, and class-conditioned generation reveal our method effectively corrects missing objects and low-quality errors.

\noindent
\textbf{Acknowledgements.}\quad
The work of Yujian Liu and Shiyu Chang was partially supported by National Science Foundation Grant IIS-2207052 and IIS-2302730. Tommi Jaakkola acknowledges support from the MIT-IBM Watson AI Lab.

%% file: sec/X_suppl.tex
\clearpage

\newcounter{alphasect}
\def\alphainsection{0}

\let\oldsection=\section
\def\section{%
  \ifnum\alphainsection=1%
    \addtocounter{alphasect}{1}
  \fi%
\oldsection}%

\renewcommand\thesection{%
  \ifnum\alphainsection=1%
    \Alph{alphasect}%
  \else
    \arabic{section}%
  \fi%
}%

\newenvironment{alphasection}{%
  \ifnum\alphainsection=1%
    \errhelp={Let other blocks end at the beginning of the next block.}
    \errmessage{Nested Alpha section not allowed}
  \fi%
  \setcounter{alphasect}{0}
  \def\alphainsection{1}
}{%
  \setcounter{alphasect}{0}
  \def\alphainsection{0}
}%

\begin{alphasection}

\section{Correctness of the Particle Filtering Framework}
\label{append:proof}

\begin{algorithm*}
\caption{Particle Filtering Framework for Correcting Diffusion Generation}
\label{alg:pf}
\begin{algorithmic}[1]
\State \textbf{Input:}
\State - Diffusion model with denosing distribution \e{q(\bm X_t | \bm X_{t+1}, \bm C)}, condition signal \e{\bm c}, number of particles \e{K}.
\State - Note: operations involving index \e{k} are performed for \e{k \in \{1, \ldots, K\}}.

\State
\State \textbf{Initialization:}
\State Sample \e{\bm x_T^{(k)} \sim q(\bm X_T)}, set \e{\phi(\bm x_T^{(k)} | \bm c) = 1}
\For{$t = T-1$ to $0$} \Comment{Iterate over time steps}
    \State \e{\tilde{\bm x}_t^{(k)} \sim q(\bm X_t | \bm x_{t+1}^{(k)}, \bm c)} \Comment{Proposal}

    \State \e{\phi(\tilde{\bm x}_t^{(k)} | \bm c) = \textsc{CalcCorrection}\left(\tilde{\bm x}_t^{(k)}, \bm c, t \right)} \Comment{Calculate correction term in Eq. \eqref{eq:likelihood_ratio}}

    \State \e{w(\bm x_{t+1}^{(k)}, \tilde{\bm x}_t^{(k)} | \bm c) = \dfrac{\phi(\tilde{\bm x}_t^{(k)} | \bm c)}{\phi(\bm x_{t+1}^{(k)} | \bm c)}} \Comment{Calculate resampling weight}
    
    \State \e{\bm x_{t}^{(k)} \sim \textsc{Multinomial}\left(\{\tilde{\bm x}_t^{(k)}\}_{k=1}^K; \{w(\bm x_{t+1}^{(k)}, \tilde{\bm x}_t^{(k)} | \bm c)\}_{k=1}^K\right)} \Comment{Resampling}
\EndFor
\State
\State \textbf{Output:}
    \State \e{\{\bm x_0^{(k)}\}_{k=1}^K} that approximately follow the ground-truth distribution \e{p(\bm X_0 | \bm c)}.
\end{algorithmic}
\end{algorithm*}

We now prove the correctness of our resampling weight design in the particle filtering algorithm. Denote the distribution that \e{\{\bm x_{t}^{(k)}\}} follows as \e{v(\bm X_t | \bm C)}, based on the particle filtering process, it is easy to show the probability satisfies the following recursive relationship:
\begin{equation}
\small
\begin{aligned}
    &v(\bm X_t | \bm C) \propto \\
    &\int v(\bm X_{t+1} | \bm C) r(\bm X_t | \bm X_{t+1}, \bm C) w(\bm X_t, \bm X_{t+1} | \bm C) \dd \bm X_{t+1},
\end{aligned}
\label{eq:particle_recursive}
\end{equation}
where \e{r(\bm X_t | \bm X_{t+1}, \bm C)} is the proposal distribution and \e{w(\bm X_t, \bm X_{t+1} | \bm C)} is the resampling weight.

Now consider \e{r(\bm X_t | \bm X_{t+1}, \bm C)=q(\bm X_t | \bm X_{t+1}, \bm C)}, \emph{i.e.,} the diffusion model is used for proposal, and \e{w(\bm X_t, \bm X_{t+1} | \bm C)=\frac{\phi(\bm X_{t} | \bm C)}{\phi(\bm X_{t+1} | \bm C)}}. Suppose \e{v(\bm X_{t+1} | \bm C) = q(\bm X_{t+1} | \bm C) \phi(\bm X_{t+1} | \bm C)}, using the recursive relationship in Eq. \eqref{eq:particle_recursive}, it can be shown that
\begin{equation}
\small
\begin{aligned}
&v(\bm X_t | \bm C) \propto \\
&\int v(\bm X_{t+1} | \bm C) q(\bm X_t | \bm X_{t+1}, \bm C) w(\bm X_t, \bm X_{t+1} | \bm C) \dd \bm X_{t+1} \\
&= \int q(\bm X_{t+1} | \bm C) \phi(\bm X_{t+1} | \bm C) q(\bm X_t | \bm X_{t+1}, \bm C) \frac{\phi(\bm X_{t} | \bm C)}{\phi(\bm X_{t+1} | \bm C)} \dd \bm X_{t+1} \\
&= \int q(\bm X_t, \bm X_{t+1} | \bm C) \phi(\bm X_{t} | \bm C) \dd \bm X_{t+1} \\
&= q(\bm X_{t} | \bm C) \phi(\bm X_{t} | \bm C).
\end{aligned}
\end{equation}
At time step \e{T}, let \e{\phi(\bm X_{T} | \bm C)=1}, and thus \e{v(\bm X_{T} | \bm C) = q(\bm X_{T} | \bm C) \phi(\bm X_{T} | \bm C)}. Therefore, by mathematical induction, \e{v(\bm X_{t} | \bm C) = q(\bm X_{t} | \bm C) \phi(\bm X_{t} | \bm C)} for all \e{t=0, \ldots, T}.

Algorithm \ref{alg:pf} describes the overall procedure of our particle filtering framework.

\section{Resampling Weight Calculation}
\label{append:resample-weight}
In this section, we provide the detailed process for calculating resampling weights, and more specifically, the correction term \e{\phi(\bm X_{t} | \bm C)}. Algorithm \ref{alg:calculate_phi} describes the overall procedure. We will mainly focus on the hybrid approach. For the discriminator-based approach, please refer to Section \ref{subsec:method1}.

\begin{algorithm*}
\caption{\textsc{CalcCorrection}}
\label{alg:calculate_phi}
\begin{algorithmic}[1]
\State \textbf{Input:}
\State - Sample \e{\tilde{\bm x}_{t}}, conditional signal \e{\bm c}, time step \e{t}.
\State - Conditional discriminator \e{d(\bm X_t | \bm C; t)}, unconditional discriminator \e{d(\bm X_t; t)}, object detector \e{\hat{p}(O_{Xi}=1 | \bm X_0)}, statistics \e{\kappa_{it}} estimated from historically generated images, hyper-parameter \e{\pi_{it}}.

\State
\If{use discriminator-based approach} \Comment{Section \ref{subsec:method1}}
\State \e{\phi(\tilde{\bm x}_t | \bm c) = \dfrac{d(\tilde{\bm x}_t | \bm c; t)}{1 - d(\tilde{\bm x}_t | \bm c; t)}} \Comment{Eq. \eqref{eq:discriminator}}
\ElsIf{use hybrid approach} \Comment{Section \ref{subsec:method2}}
\State \e{\dfrac{p(\tilde{\bm x}_t)}{q(\tilde{\bm x}_t)} = \dfrac{d(\tilde{\bm x}_t; t)}{1 - d(\tilde{\bm x}_t; t)}} \Comment{Unconditional likelihood ratio}
\State \e{p(O_{Ci}=1 | \tilde{\bm x}_{t}) \approx \hat{p}(O_{Xi}=1 | f(\tilde{\bm x}_{t}))} \Comment{Eq. \eqref{eq:numerator}}

\State \e{q(O_{Ci} = 1 | \tilde{\bm x}_{t})  \approx \hat{p}(O_{Xi} = 1 | f(\tilde{\bm x}_{t})) + \hat{p}(O_{Xi} = 0 | f(\tilde{\bm x}_{t})) \frac{(1-\kappa_{it})\pi_{it}}{(1-\kappa_{it})\pi_{it} + 1 - \pi_{it}}} \Comment{Eq. \eqref{eq:denominator}}

\State \e{\phi_t(\tilde{\bm x}_{t} | \bm c) = \dfrac{p(\tilde{\bm x}_{t})}{q (\tilde{\bm x}_{t})} \cdot \dfrac{\prod_{i: O_{Ci}=1} p(O_{Ci}=1 | \tilde{\bm x}_{t})}{\prod_{i: O_{Ci}=1} q(O_{Ci} = 1 | \tilde{\bm x}_{t})}} \Comment{Eq. \eqref{eq:hybrid-summary}}
\EndIf
\State

\State \textbf{Output:}
\State - Correction term \e{\phi_t(\tilde{\bm x}_{t} | \bm c)}.
\end{algorithmic}
\end{algorithm*}

Recall that in the hybrid approach, \e{\phi(\bm X_{t} | \bm C)} consists of two terms: unconditional likelihood ratio \e{\frac{p(\bm X_t)}{q(\bm X_t)}} and object mention ratio \e{\frac{p(\bm O_C | \bm X_t)}{q(\bm O_C | \bm X_t)}}. The unconditional likelihood ratio can be estimated with an unconditional discriminator as in Eq.\eqref{eq:discriminator}. We will next elaborate on how to estimate the object mention ratio.

\vspace{0.05in}
\noindent
\textbf{Estimating object mention ratio.}\quad
As shown in Eq. \eqref{eq:object_occur_decomp}, the object mention ratio can be decomposed as
\begin{align*}
\small
\frac{\prod_{i: O_{Ci}=1} p(O_{Ci}=1 | \bm X_t)}{\prod_{i: O_{Ci}=1} q(O_{Ci} = 1 | \bm X_t)} \cdot \frac{\prod_{i: O_{Ci}=0} p(O_{Ci}=0 | \bm X_t)}{\prod_{i: O_{Ci}=0} q(O_{Ci} = 0 | \bm X_t)},
\end{align*}
where we focus on the first term that corresponds to the missing object errors.

To compute \e{p(O_{Ci}=1 | \bm X_t)}, notice that under \e{p} distribution, \e{O_{Ci} = O_{Xi}}, because, by definition, a text caption is constructed to mention the objects that appear in the corresponding real image. Therefore,
\begin{equation}
\small
\begin{aligned}
    p(O_{Ci}=1 | \bm X_t) &= p(O_{Xi}=1 | \bm X_t)\\ &\stackrel{\text{\ding{172}}}{=} \mathbb{E}_{\bm X_0 \sim p(\bm X_0 | \bm X_t)}[p(O_{Xi } = 1| \bm X_0)]\\
    &\stackrel{\text{\ding{173}}}{\approx} \hat{p}(O_{Xi } = 1 | f(\bm X_t)),
\end{aligned}
\label{eq:p_occur}
\end{equation}
where equality \ding{172} is by simple chain rule and the fact that \e{p(O_{Xi }| \bm X_0, \bm X_t) = p(O_{Xi }| \bm X_0)} because \e{O_{Xi}} is defined as the object occurrence in the clean image \e{\bm X_0}, and hence is conditionally independent of \e{\bm X_t}. For \ding{173}, we make two approximations. First, we replace drawing a sample of \e{\bm X_0} from \e{p(\bm X_0 | \bm X_t)} with the minimum mean squared error (MMSE) estimate of \e{\bm X_0} from \e{\bm X_t}, denoted as \e{f(\bm X_t)}, which can be conveniently and efficiently predicted by the denoising network via one-step generation of \e{\bm X_0} from \e{\bm X_t}. Second, we approximate the true object occurrence probability \e{p(O_{Xi} | \bm X_0)} with the one estimated by the object detector \e{\hat{p}(O_{Xi} | \bm X_0)}.

On the other hand, \e{q(O_{Ci}=1 | \bm X_t)} can be most intuitively computed by training a neural network to predict object mentions in the input text from the generated samples. However, this approach is inefficient and would need to re-train the network everytime the denoising scheme changes. We would like to derive a more efficient alternative taking advantage of the object detector. Formally (notice that it is no longer true that \e{O_{Ci} = O_{Xi}}),
\begin{equation}
\small
\begin{aligned}
q(O_{Ci}=1 | \bm X_t) &= \mathbb{E}_{O_{Xit}\sim q(O_{Xit} | \bm X_t)} [q(O_{Ci}=1 | O_{Xit}, \bm X_t )] \\
&= q(O_{Xit}=0 | \bm X_t) q(O_{Ci}=1 | O_{Xit}=0, \bm X_t) + \\
&q(O_{Xit}=1 | \bm X_t) q(O_{Ci}=1 | O_{Xit}=1, \bm X_t),
\end{aligned}
\label{eq:q_occur}
\end{equation}
where \e{O_{Xit}} denotes the occurrence of object \e{i} in the clean image predicted from \e{\bm X_t}, \emph{i.e.,} \e{f(\bm X_t)}.
Similar to Eq.~\eqref{eq:p_occur}, the probability \e{q(O_{Xit} | \bm X_t)} can be estimated by \e{\hat{p}(O_{Xi } | f(\bm X_t))}.Therefore, what makes Eq. \eqref{eq:q_occur} different is the additional term \e{q(O_{Xit}=0 | \bm X_t) q(O_{Ci}=1 | O_{Xit}=0, \bm X_t)}, which corresponds to the case where even if the object detector fails to detect the object \e{i}, there is still chance that the caption used to generate the image contains object \e{i}, considering the imperfect diffusion model that may miss objects mentioned in the caption.

To calculate \e{q(O_{Ci}=1 | O_{Xit}, \bm X_t)}, we further make an assumption that object occurrence \e{O_{Xit}} is a sufficient statistic to predict the object mention of the same object. Hence \e{q(O_{Ci}=1 | O_{Xit}, \bm X_t) = q(O_{Ci}=1 | O_{Xit})}. According to the Bayes rule,
\begin{equation}
    \small
    q(O_{Ci}=1 | O_{Xit}) = \frac{q(O_{Xit} | O_{Ci}=1)q(O_{Ci}=1)}{\sum_{O_{Ci}\in \{0, 1\}}q(O_{Xit} | O_{Ci})q(O_{Ci})}.
\end{equation}
Both \e{q(O_{Xit} | O_{Ci}=1)} and \e{q(O_{Xit} | O_{Ci}=0)} can be estimated from the samples of distribution \e{q(\bm X_t | \bm C)}. Specifically, consider a different set of \e{H} samples \e{\{\bm x_t^{(h)} \}} that are generated without particle filtering process. For each sample \e{\bm x_t^{(h)}}, we compute the object occurrence estimate \e{\hat{O}_{Xit}} by feeding the predicted clean image, \e{f(\bm x_t^{(h)})}, to the object detector, and thresholding the output probability of the object detector with \e{0.5}. Then \e{q(O_{Xit}=1 | O_{Ci}=1)} can be estimated by calculating the percentage of samples whose \e{\hat{O}_{Xit}=1} among the samples whose corresponding input caption mentions object \e{i}, as in Eq. \eqref{eq:kappa}. \e{q(O_{Xit}=1 | O_{Ci}=0)} can be calculated similarly from samples whose input caption does not mention object \e{i}. In practice, since the diffusion model seldom generates objects that are not mentioned in the text, we empirically find that \e{q(O_{Xit} = 1 | O_{Ci}=0)} is very close to zero. Hence, to reduce computation complexity, we will set this term to zero, and \e{q(O_{Xit} = 0 | O_{Ci}=0)} to one. Finally, \e{q(O_{Ci}=0)} and \e{q(O_{Ci}=1)} controls how much weight we would like to place on each case, and we will treat them as hyper-parameters and denote \e{q(O_{Ci}=1)} as \e{\pi_{it}}. Note that although \e{\pi_{it}} can be dependent on both \e{i} and \e{t}, for simplicity, we set \e{\pi_{it}} to be the same for all \e{i} and \e{t}.

Denote \e{q(O_{Xit}=1 | O_{Ci}=1)} as \e{\kappa_{it}}, with the above assumptions,
\begin{equation}
\small
\begin{aligned}
    &q(O_{Ci}=1 | O_{Xit}=1) = 1 \\
    &q(O_{Ci}=1 | O_{Xit}=0) = \frac{(1-\kappa_{it})\pi_{it}}{(1-\kappa_{it})\pi_{it} + 1 - \pi_{it}}.
\end{aligned}
\label{eq:q_mention_occur}
\end{equation}
Plug Eq. \eqref{eq:q_mention_occur} back to Eq. \eqref{eq:q_occur}, we can derive Eq. \eqref{eq:denominator} in the main paper.

\section{Experiments on Text-to-image Generation}
\label{append:t2i}

\subsection{Reparameterize Stable Diffusion for Reverse-time SDE}
Following \citet{karras2022elucidating} and \citet{xu2023restart}, we use the reverse-time SDE \cite{Anderson1982ReversetimeDE} for generation:
\begin{equation}
\small
\dd \bm x = -2\dot{\sigma}(t)\sigma(t)\nabla_{\bm x}\log p(\bm x; \sigma(t)) \dd t + \sqrt{2\dot{\sigma}(t)\sigma(t)} \dd \omega_t,
\label{eq:probability-ode}
\end{equation}
where \e{\omega_t} is the standard Wiener process, \e{\sigma(t) = t} is the noise level at time \e{t}, and \e{p(\bm x; \sigma(t))} is the distribution obtained by adding Gaussian noise of standard deviate \e{\sigma(t)} to the input data.

To use stable diffusion \citep{rombach2022high} in Eq. \eqref{eq:probability-ode}, the key is to reparameterize the model to estimate the score \e{\nabla_{\bm x}\log p(\bm x; \sigma(t))}. Following \citet{karras2022elucidating}, we estimate the score as follows:
\begin{equation}
\small
\begin{aligned}
    &\nabla_{\bm x}\log p(\bm x; \sigma) = \frac{D(\bm x; \sigma) - \bm x}{\sigma^2}, \\
    &D(\bm x; \sigma) = c_{\mathrm{skip}}(\sigma) \bm x + c_{\mathrm{out}}(\sigma)F\left(c_{\mathrm{in}}(\sigma)\bm x; c_{\mathrm{noise}}(\sigma)\right),
\end{aligned}
\end{equation}
where \e{F(\cdot)} is the denoising network trained in stable diffusion, \e{c_{\mathrm{skip}}(\sigma) = 1}, \e{c_{\mathrm{out}}(\sigma) = -\sigma}, \e{c_{\mathrm{in}}(\sigma) = 1/\sqrt{\sigma^2+1}}, \e{c_{\mathrm{noise}}(\sigma) = 999\sigma_{\mathrm{train}}^{-1}(\sigma)}, \e{\sigma_{\mathrm{train}}(t)} is the noise schedule used when training the denoising network, and \e{\sigma_{\mathrm{train}}^{-1}(\sigma)} denotes the inverse of \e{\sigma_{\mathrm{train}}(t)}.
In particular, the forward process of Stable Diffusion v2.1-base can be considered as a discrete version of the VP SDE \citep{song2021scorebased}:
\begin{equation}
\small
\dd \bm x = f(t)\bm x \dd t + g(t) \dd \omega_t,
\label{eq:sd_forward}
\end{equation}
where
\begin{equation}
\small
\begin{aligned}
    &f(t)=-\frac{1}{2}\beta(t),\quad g(t)=\sqrt{\beta(t)},\\
    &\beta(t) = \left( (\beta_{\textrm{max}}^{0.5}-\beta_{\textrm{min}}^{0.5})t + \beta_{\textrm{min}}^{0.5} \right)^2,
\end{aligned}
\end{equation}
with \e{\beta_{\textrm{min}} = 0.85}, and \e{\beta_{\textrm{max}} = 12}.
We follow \citet{karras2022elucidating} (Eqs. (152) - (163)) to derive \e{\sigma_{\mathrm{train}}(t)} by plugging in the scaled linear schedule \e{\beta(t) = \left( (\beta_{\textrm{max}}^{0.5}-\beta_{\textrm{min}}^{0.5})t + \beta_{\textrm{min}}^{0.5} \right)^2} used in Stable Diffusion v2.1-base, which results in:
\begin{equation}
\small
\sigma_{\mathrm{train}}(t) = \sqrt{e^{\frac{1}{3}\beta_d^2 t^3 + \beta_d \beta_{\textrm{min}}^{0.5} t^2 + \beta_{\textrm{min}}t} - 1},
\label{eq:sd_sigma}
\end{equation}
where \e{\beta_d = \beta_{\textrm{max}}^{0.5}-\beta_{\textrm{min}}^{0.5}}.

Finally, although Stable Diffusion is trained with discrete time steps, we follow \citet{lu2022dpmsolver} to directly feed the continuous time value \e{c_{\mathrm{noise}}(\sigma)} to the model.

\subsection{Implementation Details for Baselines}
\label{append:implement-details}
\textbf{Identify objects in the caption.}\quad
For methods that require the identification of objects in the caption, we use the following two steps:
\begin{enumerate}
    \item Extract all noun phrases in the caption using spaCy.\footnote{\url{https://spacy.io}.} Filter out noun phrases that are stop words, \emph{e.g.,} ``which'' and ``who''.
    \item Check if extracted noun phrases match one of the object categories in \texttt{MS-COCO}, and only keep the phrases that belong to \texttt{MS-COCO} objects. The match is determined by whether the Levenshtein Distance between the noun phrase and object category name is smaller than 0.1 of the length of the noun phrase and object category name.
\end{enumerate}
\noindent
The same process is also used to filter out 261 captions in \texttt{MS-COCO} that contain at least four objects (to measure object occurrence), and 5,000 captions in \texttt{MS-COCO} that contain at least one object (to measure FID).

\vspace{0.05in}
\noindent
\textbf{Sample selection criteria.}\quad
For \textsc{Objectselect}, we use Eq. \eqref{eq:simple-selection-criterion} to select the best image from generated images. Specifically, since the object detector outputs multiple proposal regions, where each proposal region comes with a predicted probability for object \e{i}, we take the maximum probability for object \e{i} over all proposal regions as \e{\hat{p}(O_{Xi } = 1 | \bm x_0)}. For \textsc{Rewardselect} and \textsc{TIFAselect}, we use the official implementation \citep{karthik2023dont} to select images, except that we only include object-related questions for \textsc{TIFAselect} since our focus is on object occurrence.

For other baselines, we use the official implementation to generate images. For \textsc{Spatial-temporal}, we only optimize for five rounds since we found further optimization does not improve performance. For \textsc{D-guidance}, we only apply guidance when \e{\sigma(t) < 5} since we found at large noise levels, the object detector barely detects any object, which results in a noisy gradient.
All methods are evaluated on both Restart sampler \citep{xu2023restart} and EDM sampler \citep{karras2022elucidating}. The only exceptions are \textsc{Spatial-temporal} and \textsc{Attend-excite}, which use the original sampler in their papers.

\subsection{Discriminator Training}
We follow \citet{kim2023refining} to train both unconditional and conditional discriminators. Table \ref{tab:discriminator_t2i} shows the hyper-parameters for discriminator training.

\vspace{0.05in}
\noindent
\textbf{Training data and objective.}\quad
We use LAION Aesthetics \citep{schuhmann2022laion5b} to train discriminators. We randomly sample 1M images with aesthetics score higher than 6 as real images. For fake images, we generate 1M images with 49 NFE using EDM sampler. For conditional discriminator, the captions used to generate fake images are the same with the captions of the real images. For unconditional discriminator, we use a different set of 1M captions to generate fake images.

The discriminators are trained with the canonical discrimination loss in Eq. \eqref{eq:loss}. Specifically, given a clean image, we uniformly sample time \e{t \in [10^{-5}, 1]} and follow Eq. \eqref{eq:sd_forward} to get the corresponding noisy image. The discriminator is then trained to discriminate noisy version of real and fake images.

\begin{table}
    \centering
    \resizebox{\linewidth}{!}{
    \begin{tabular}{lcc}
    \topline
     &  \textbf{Unconditional} & \textbf{Conditional} \\
     \midtopline
     \# real images& 1,000,000 & 1,000,000 \\
     \# fake images& 1,000,000 & 1,000,000 \\
     Same captions for real \& fake images? & \ding{56} & \ding{52} \\
     \# Epochs & 2 & 2 \\
     Batch size & 128 & 128 \\
     Learning rate & $10^{-4}$ & $10^{-4}$ \\
     Time sampling & Uniform & Uniform \\
     Total GPU hours & 22 & 22 \\
     GPUs & A6000 & A6000 \\
     \bottomline
    \end{tabular}
    }
    \caption{Configurations of discriminator training for text-to-image generation.}
    \vspace{-2mm}
    \label{tab:discriminator_t2i}
\end{table}

\vspace{0.05in}
\noindent
\textbf{Model architecture.}
We use the hidden representations of the middle block of the U-Net \citep{ronneberger2015unet} to predict whether an image is real or fake, since previous works have found that these representations capture semantic information in the input image \citep{kwon2023diffusion, baranchuk2022labelefficient}. Specifically, we initialize the discriminator with the U-Net in Stable Diffusion v2.1-base and only keep its down and middle blocks. We add a linear prediction layer on top the representations of the middle block. During training, we freeze the down block and only fine-tune the middle block and the prediction layer.

\subsection{Sampling Configurations}
\label{append:t2i-sampling-config}
For all sample selection methods (including ours), we generate each image with a fixed NFE and report performance when \e{K=5, 10, 15} images are generated.

Table \ref{tab:t2i_config} shows the sampling configurations for both samplers. For Restart sampler, \e{N_{\mathrm{main}}} denotes the number of steps in the main backward process, \e{N_{\mathrm{Restart}, i}, K_i, t_{\mathrm{min}, i}, t_{\mathrm{max}, i}} is the number of steps, number of repetitions, minimal, and maximum time points in the \e{i}-th restart interval, and \e{l} is the number of restart intervals. Following \citet{xu2023restart}, we use Euler method for the main backward process and Heun method for the restart backward process. Please refer to the original paper for detailed explanations. For our \textsc{PF} methods, the resampling is always performed before adding noise, \emph{i.e.,} at time \e{t_{\mathrm{min}, i}}, thus resampling is perfromed six times in total. For EDM sampler, \e{N} denotes the number of denoising steps, \e{n_i} denotes the index of steps when resampling is performed (for \textsc{PF} methods), and \e{m} denotes the number of resampling steps. Resampling is only performed four times because we empirically find performing resampling at each denoising step does not improve performance significantly compared to resampling only at a subset of steps.

Table \ref{tab:computation_cost} shows the comparison of computation cost for all methods. The reported cost for sample selection methods is when \e{K=5} images are generated for each caption, since their performance already exceeds other baselines at \e{K=5}. Based on the table, all methods are significantly faster than \textsc{Spatial-temporal}, and sample selection methods have similar runtime with \textsc{Attend-excite} on EDM sampler and \textsc{D-guidance} on Restart sampler. Moreover, the sample selection methods can be run in parallel much easier than other methods, so their runtime can potentially be further reduced.

\begin{table}
    \centering
    \resizebox{\linewidth}{!}{
    \begin{tabular}{c}
    \topline
    \textbf{Restart configuration} \\
    \e{N_{\mathrm{main}}, \{(N_{\mathrm{Restart}, i}, K_i, t_{\mathrm{min}, i}, t_{\mathrm{max}, i})\}_{i=1}^l} \\
     \midrule
     \e{30, \{(4, 1, 1.09, 1.92), (4, 2, 0.59, 1.09), (4, 2, 0.30, 0.59), (4, 1, 0.06, 0.30)\} } \\
     \midrule
     \textbf{EDM configuration}: \e{N, \{n_i\}_{i=1}^m} \\
     \midrule
     \e{25, \{10, 13, 16, 19\}} \\
     \bottomline
    \end{tabular}
    }
    \caption{Sampling configurations for text-to-image generation.}
    \label{tab:t2i_config}
\end{table}

\begin{table*}
\centering
\resizebox{0.85\linewidth}{!}{
\begin{tabular}{lcccccc}
    \topline
     \multicolumn{1}{c}{\textbf{}} & \multicolumn{3}{c}{\textbf{EDM Sampler}} & \multicolumn{3}{c}{\textbf{Restart Sampler}} \\
\cmidrule(lr){2-4}
\cmidrule(lr){5-7}
     & NFE & Require Gradient? & Runtime (s) & NFE & Require Gradient? & Runtime (s) \\
    \midtopline
     \textsc{SD} \citep{rombach2022high} & 49 & \ding{56} & 7 & 66 & \ding{56} & 9 \\
     \textsc{D-guidance} \citep{kim2023refining} & 67 & \ding{52} & 20 & 106 & \ding{52} & 41 \\
     \textsc{Spatial-temporal}$^*$ \citep{wu2023harnessing} & 250 & \ding{52} & 110 & -- & -- & -- \\
     \textsc{Attend-excite}$^*$ \citep{chefer2023attendandexcite} & 50 & \ding{52} & 29 & -- & -- & -- \\
     \textsc{TIFAselect} \citep{karthik2023dont} & 245 & \ding{56} & 39 & 330 & \ding{56} & 50 \\
     \textsc{Rewardselect} \citep{karthik2023dont} & 245 & \ding{56} & 32 & 330 & \ding{56} & 43 \\
     \textsc{Objectselect} & 245 & \ding{56} & 32 & 330 & \ding{56} & 43 \\
     \textsc{PF-discriminator} & 245 & \ding{56} & 33 & 330 & \ding{56} & 44 \\
     \textsc{PF-hybrid} & 265 & \ding{56} & 39 & 360 & \ding{56} & 53 \\
    \bottomline
\end{tabular}
}
\caption{Computation cost for all methods. Runtime is measured on a single NVIDIA V100 GPU. $^*$: the two baselines use the original samplers in their papers.}
\label{tab:computation_cost}
\end{table*}

\begin{figure*}
  \centering
  \begin{subfigure}{0.8\linewidth}
    \includegraphics[width=\linewidth]{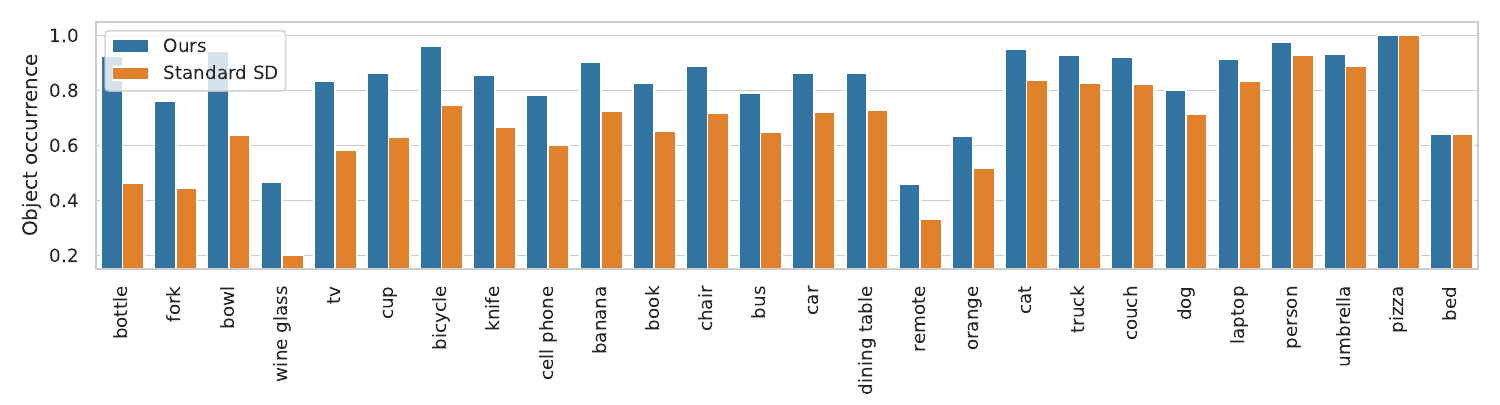}
    \vspace{-5mm}
    \caption{Class-wise object occurrence of our method and standard \textsc{SD} on \texttt{MS-COCO}.}
    \label{fig:class-obj-occur}
  \end{subfigure}
  \begin{subfigure}{0.8\linewidth}
    \includegraphics[width=\linewidth]{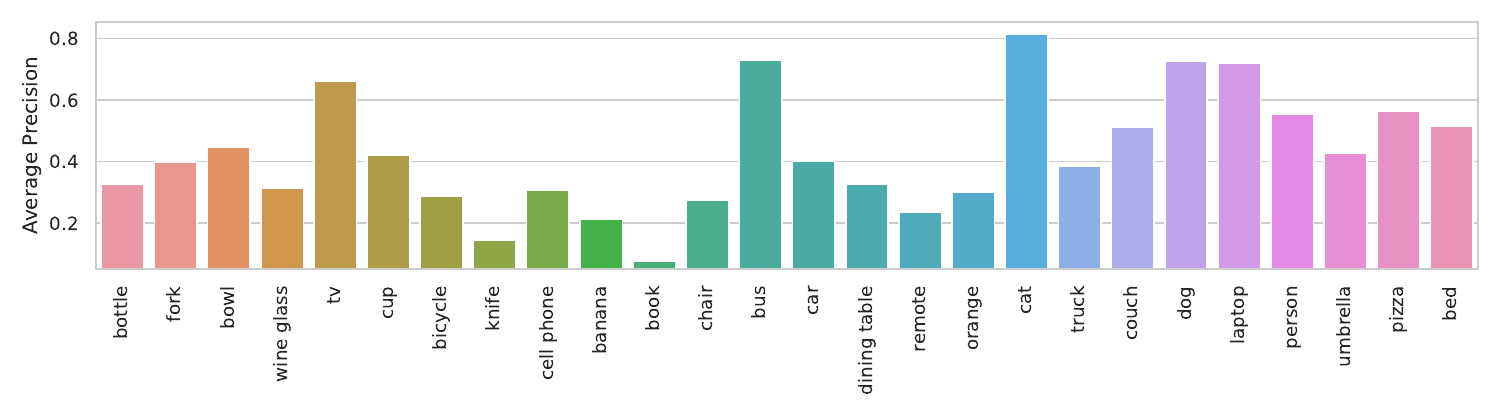}
    \vspace{-5mm}
    \caption{Class-wise average precision of the object detector on \texttt{MS-COCO}.}
    \label{fig:class-ap}
  \end{subfigure}
  \vspace{-2mm}
  \caption{Class-wise performance of our method and the object detector.}
  \label{fig:class-perf}
\end{figure*}

\subsection{Performance Breakdown by Object Category}
\label{append:object-category}
In this section, we show the performance of our method for each object category in order to study \ding{182} Which objects benefit more from our method; and \ding{183} How does the object detector affect our method. Figure \ref{fig:class-obj-occur} shows the class-wise object occurrence of our method and standard \textsc{SD} on \texttt{MS-COCO}, where classes are sorted by the amount of improvement achieved by our method and we only keep the classes that appear in at least 5 captions. As can be observed, our method improves or matches the performance of the standard SD on all objects. It is particularly beneficial on small objects with fine details, such as bottles, forks, and wine glasses. Figure \ref{fig:class-ap} further shows the class-wise average precision of the object detector used during generation, evaluated on \texttt{MS-COCO}. As shown in the figure, the performance of our method does not correlate well with the performance of the detector. Instead, it depends more on the capability of the diffusion model in correctly generating the objects.

\subsection{Additional Results with EDM Sampler}
\label{append:t2i-edm-results}

\begin{figure}
    \centering
    \includegraphics[width=\linewidth]{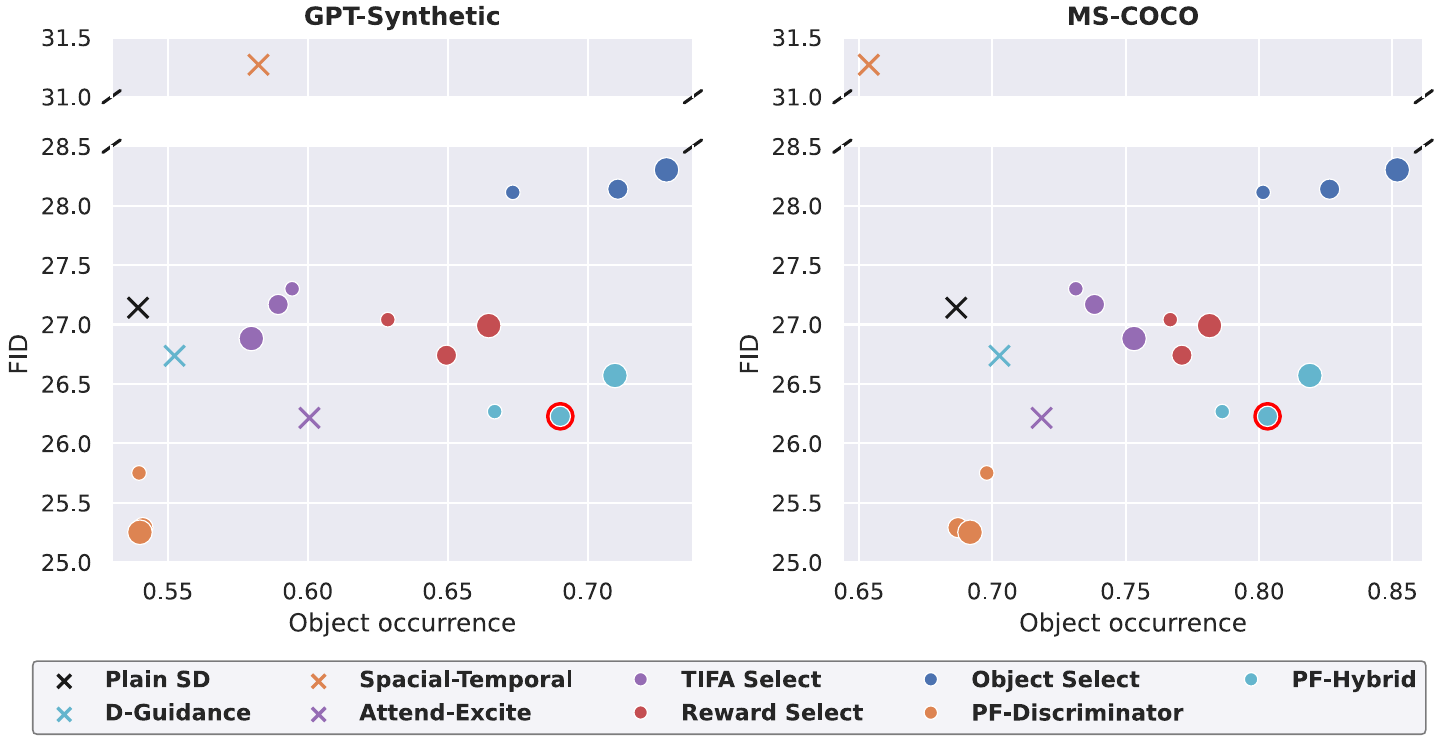}
    \caption{FID ($\downarrow$) \emph{vs.} Object occurrence ($\uparrow$) for all methods evaluated with EDM sampler. Ideal points should scatter at the bottom right corner. Object occurrence is measured on \texttt{GPT-Synthetic} (left) and \texttt{MS-COCO} (right), and FID is measured on \texttt{MS-COCO}. \e{K=5, 10, 15} images are generated for sample selection methods, and the sizes of points indicate the value of \e{K} (larger \e{K} has larger points). The method that achieves the best combined performance is highlighted in \textcolor{red}{red}.}
    \label{fig:t2i-edm-result}
\end{figure}

Figure \ref{fig:t2i-edm-result} shows the results when EDM is used as the underlying sampler. There are two observations from the figure. First, the overall trend of all methods aligns with Figure \ref{fig:t2i-results} when Restart is used as the sampler. In particular, the sampling-based methods generally outperform the non-sampling-based ones. And the three methods proposed in this paper, \textsc{Objectselect}, \textsc{PF-discriminator}, and \textsc{PF-hybrid} lie at the frontier of the performance trade-off, significantly outperform other methods. Furthermore, \textsc{PF-hybrid} is the only method that simultaneously achieves a high object occurrence and a low FID. Second, the performance of EDM sampler is generally worse than that of Restart sampler on both object occurrence and FID, which shows the superiority of the restart process.

\begin{figure}
    \centering
    \includegraphics[width=0.7\linewidth]{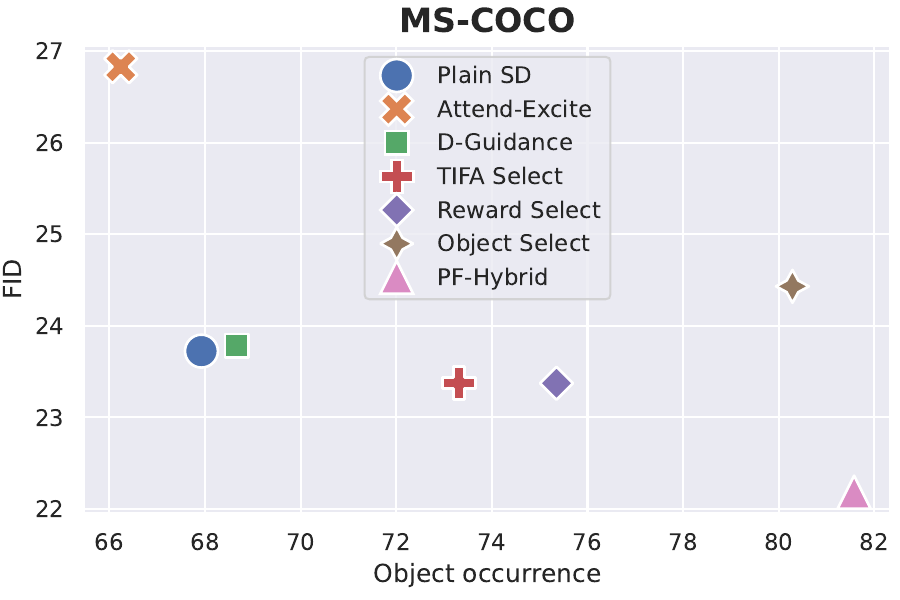}
    \caption{FID ($\downarrow$) \emph{vs.} Object occurrence ($\uparrow$) for all methods evaluated with Stable Diffusion v1.5 and Restart sampler. Ideal points should scatter at the bottom right corner. Object occurrence and FID are both measured on \texttt{MS-COCO}. \e{K=10} images are generated for sample selection methods.}
    \label{fig:t2i-sd15-result}
\end{figure}

\subsection{Additional Results with SD 1.5}
\label{append:t2i-sd15-results}
We also experiment with Stable Diffusion v1.5 on \texttt{MS-COCO}. The results are shown in Figure \ref{fig:t2i-sd15-result}. We observe a similar trend with the results from Stable Diffusion v2.1-base (Figure \ref{fig:t2i-results}), where our method achieves the best overall performance among baselines.

\section{Experiments on Unconditional and Class-conditioned Generation}
\label{append:benchmark}

\subsection{Discriminator Training}
\label{append:benchmark-discriminator-training}
We follow the official implementation \citep{kim2023refining} to train discriminators on \texttt{ImageNet-64} and \texttt{FFHQ} since the original paper did not experiment on the dataset or did not release the trained discriminator. Table \ref{tab:discriminator_benchmark} shows the training configurations. Following \citet{kim2023refining}, the discriminators use features extracted by the frozen pre-trained classifier in \citet{dhariwal2021diffusion}, and a shallow U-Net is added to discriminate images. During training, the pre-trained classifier is frozen and only the shallow U-Net is trained. Please refer to \citet{kim2023refining} for more details.

\begin{table}
    \centering
    \resizebox{0.85\linewidth}{!}{
    \begin{tabular}{lcc}
    \topline
     &  \textbf{\texttt{ImageNet-64}} & \textbf{\texttt{FFHQ}} \\
     \midtopline
     \multicolumn{3}{c}{\textbf{Training Configurations}} \\
     \midrule
     \# real images& 1,200,000 & 60,000 \\
     \# fake images& 1,200,000 & 60,000 \\
     \# Epochs & 50 & 50 \\
     Batch size & 1024 & 256 \\
     Learning rate & $5\times10^{-4}$ & $3\times10^{-4}$ \\
     Time sampling & Importance & Uniform \\
     Total GPU hours & 70 & 4 \\
     GPUs & A6000 & A6000 \\
     \midrule
     \multicolumn{3}{c}{\textbf{Shallow U-Net Architecture}} \\
     \midrule
     Input shape & \e{(B, 8, 8, 512)} & \e{(B, 8, 8, 512)} \\
     Class condition & \ding{52} & \ding{56} \\
     \# Resnet blocks & 4 & 4 \\
     \# Attention blocks & 3 & 3 \\
     \bottomline
    \end{tabular}
    }
    \caption{Configurations of discriminator training for class-conditioned (\texttt{ImageNet-64}) and unconditional (\texttt{FFHQ}) generation.}
    \label{tab:discriminator_benchmark}
\end{table}

\subsection{Sampling Configurations}
In order to have a fair comparison in terms of computation cost, we compare all methods under the same total NFE. For \textsc{PF} and \textsc{D-select}, we generate each image with a fixed NFE and vary the number of particles in generation, \emph{i.e.,} choose value of \e{K} from \e{\{2, 4, 6\}}. For the original sampler and \textsc{D-guidance}, we increase the number of steps or restart iterations to match the total NFE of sampling-based methods. All methods are evaluated on both Restart and EDM samplers. Table \ref{tab:edm_config} and \ref{tab:restart_config} show the sampling configurations for each sampler.

For EDM sampler, we use the same hyper-parameters as the original paper, except the \e{S_{\mathrm{churn}}} (which controls the amount of noise added in each denoising step) on \texttt{FFHQ}, since we find adding small amount of noise (\e{S_{\mathrm{churn}} > 0}) outperforms the original setting where no noise is added (\e{S_{\mathrm{churn}} = 0}). Additionally, for time steps after resampling is performed, we increase \e{S_{\mathrm{churn}}} to \e{(\sqrt{2}-1)N} (\e{N} is the number of denoising steps), which is the maximum allowed value in \citet{karras2022elucidating}. Since we find a large added noise after resampling can improve performance (please see Section \ref{append:ablation}). For Restart sampler, we use the same configurations as the original paper on \texttt{ImageNet-64} and slightly modify them to adapt to \texttt{FFHQ} since the original paper did not experiment on \texttt{FFHQ}.

\begin{table}
    \centering
    \resizebox{0.8\linewidth}{!}{
    \begin{tabular}{ccc}
    \topline
    \textbf{NFE per Image} & \textbf{Resampling Steps}: \e{\{n_i\}_{i=1}^m} & \e{S_{\mathrm{churn}}} \\
     \midtopline
     \multicolumn{3}{c}{\textbf{\texttt{ImageNet-64}}} \\
     \midrule
     \rowcolor{gray!20}
     127 & $\{31, 35, 39, 43, 47, 51\}$ & 10 \\
     255 & -- & 20 \\
     511 & -- & 40 \\
     767 & -- & 60 \\
     \midrule
     \multicolumn{3}{c}{\textbf{\texttt{FFHQ}}} \\
     \midrule
     \rowcolor{gray!20}
     63 & $\{15, 19, 23, 27\}$ & 1.25 \\
     127 & -- & 2.5 \\
     255 & -- & 5.0 \\
     399 & -- & 7.8125 \\
     \bottomline
    \end{tabular}
    }
    \caption{Sampling configurations for EDM sampler on standard benchmarks. Two highlighted rows are used in \textsc{PF} and \textsc{D-select} to generate one image. The resampling steps are only applicable to our \textsc{PF} methods. \e{S_{\mathrm{churn}}} controls the amount of noise added in each denoising step (please refer to \citet{karras2022elucidating} for details).}
    \label{tab:edm_config}
\end{table}

\begin{table*}
    \centering
    \resizebox{0.85\linewidth}{!}{
    \begin{tabular}{cc}
    \topline
    \multirow{2}{*}{\textbf{NFE per Image}} & \textbf{Restart Configuration} \\
    & \e{N_{\mathrm{main}}, \{(N_{\mathrm{Restart}, i}, K_i, t_{\mathrm{min}, i}, t_{\mathrm{max}, i})\}_{i=1}^l} \\
     \midtopline
     \multicolumn{2}{c}{\textbf{\texttt{ImageNet-64}}} \\
     \midrule
     67 & \e{18, \{(5, 1, 19.35, 40.79),(5, 1, 1.09, 1.92), (5, 1, 0.59, 1.09), (5, 1, 0.06, 0.30)\} } \\
     \rowcolor{gray!20}
     99 & \e{18, \{(3, 1, 19.35, 40.79),(4, 1, 1.09, 1.92), (4, 4, 0.59, 1.09), (4, 1, 0.30, 0.59), (4, 4, 0.06, 0.30)\}} \\
     165 & \e{18, \{(3, 1, 19.35, 40.79),(4, 1, 1.09, 1.92), (4, 5, 0.59, 1.09), (4, 5, 0.30, 0.59), (4, 10, 0.06, 0.30)\}}\\
     203 & \e{36, \{(4, 1, 19.35, 40.79),(4, 1, 1.09, 1.92), (4, 5, 0.59, 1.09), (4, 5, 0.30, 0.59), (6, 6, 0.06, 0.30)\}}\\
     385 & \e{36, \{(3, 1, 19.35, 40.79),(6, 1, 1.09, 1.92), (6, 5, 0.59, 1.09), (6, 5, 0.30, 0.59), (6, 20, 0.06, 0.30)\}}\\
     535 & \e{36, \{(6, 1, 19.35, 40.79),(6, 1, 1.09, 1.92), (7, 6, 0.59, 1.09), (7, 6, 0.30, 0.59), (7, 25, 0.06, 0.30)\}}\\
     \midrule
     \multicolumn{2}{c}{\textbf{\texttt{FFHQ}}} \\
     \midrule
     \rowcolor{gray!20}
     67 & \e{18, \{(5, 1, 1.09, 1.92), (5, 2, 0.59, 1.09), (5, 1, 0.06, 0.30)\}} \\
     119 & \e{18, \{(8, 1, 19.35, 40.79), (8, 2, 1.09, 1.92), (8, 2, 0.59, 1.09), (8, 1, 0.06, 0.30)\}} \\
     251 & \e{36, \{(11, 1, 19.35, 40.79), (11, 3, 1.09, 1.92), (11, 3, 0.59, 1.09), (11, 2, 0.06, 0.30)\}} \\
     401 & \e{48, \{(18, 1, 19.35, 40.79), (18, 3, 1.09, 1.92), (18, 3, 0.59, 1.09), (18, 2, 0.06, 0.30)\}} \\
     \bottomline
    \end{tabular}
    }
    \caption{Sampling configurations for Restart sampler on standard benchmarks. Two highlighted rows are used in \textsc{PF} and \textsc{D-select} to generate one image.}
    \label{tab:restart_config}
\end{table*}

\subsection{Additional Results with EDM Sampler}
\label{append:benchmark-edm-results}
Figure \ref{fig:edm_fid} shows FID as a function of total NFE when EDM is used as the sampler. Additionally, we also plot \textsc{PF} with Restart sampler for reference. There are three observations from the figure. First, FIDs of all methods generally decrease as NFE increases, and \textsc{D-guidance}, \textsc{D-select}, and \textsc{PF} all outperform the original sampler. Second, our \textsc{PF} with Restart sampler achieves the lowest FID on both datasets. Third, the performance gain of our method when combined with EDM sampler is not as large as that when combined with Restart sampler. The difference can be ascribed to the fact that resampling is performed at smaller time steps for Restart sampler, which better selects a promising population among the particles since the discriminator can better distinguish real and model-generated images at smaller time steps. Additionally, the added noise after resampling is larger for Restart, which also benefits the exploration around the promising particles (see Section \ref{append:ablation} for the impact of amount of added noise).

\begin{figure}
    \centering
    \includegraphics[width=\linewidth]{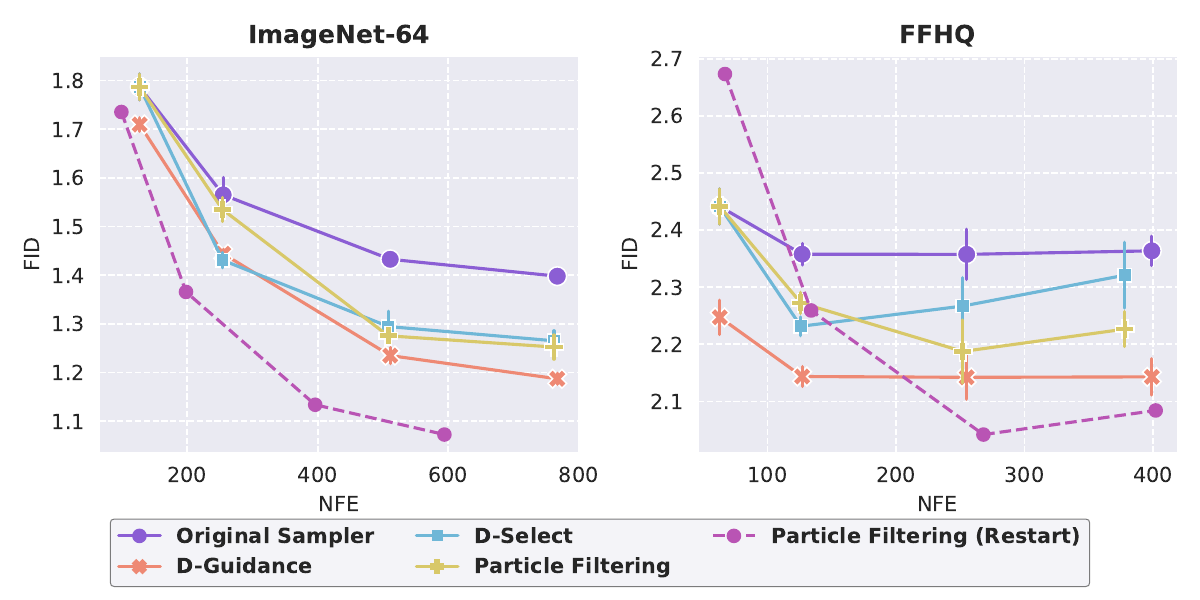}
    \caption{FID (average of 3 runs) on \texttt{ImageNet-64} (left) and \texttt{FFHQ} (right) when evaluated with EDM sampler. The \textsc{PF} with Restart sampler (dashed line) is added for reference. Error bars indicate standard deviations.}
    \label{fig:edm_fid}
\end{figure}

\subsection{Additional Results with Effective NFE}
\label{append:benchmark-effectivenfe}
As mentioned in Section \ref{subsec:exp-unconditional}, NFE is not an adequate measure of computation cost in our setting, since it only measures the number of evaluations of the denoising U-Net and ignores other compute costs such as the forward and backward passes of the discriminator. If all these costs are considered, \textsc{D-Guidance} incurs 1.55$\times$ compute cost per NFE compared to the original sampler, whereas our method only incurs 1.02$\times$ compute cost per NFE compared to the original sampler. To have a fair comparison, we re-plot FID against \textit{effective NFE}, which corrects for the disparage in compute costs per NFE. We further add a new point for \textsc{D-Guidance} at small NFE and remove a point at excessively large NFE for better comparison. The results shown in Figure \ref{fig:restart_fid_effectivenfe} illustrates that our method achieves the best performance across all compute costs

\begin{figure}
    \centering
    \includegraphics[width=\linewidth]{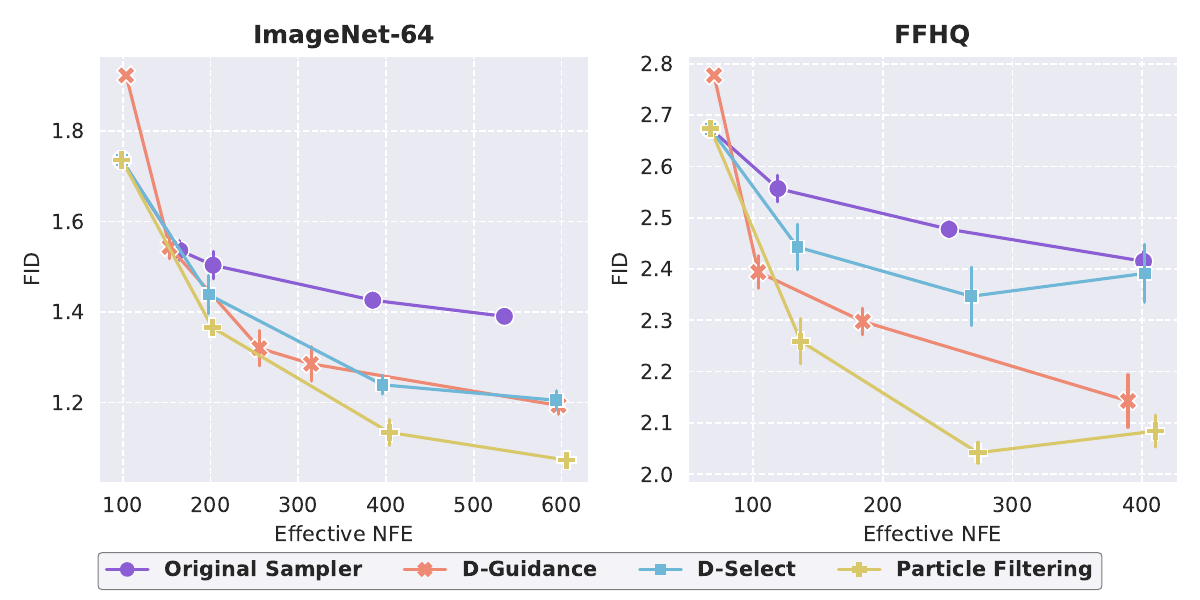}
    \caption{FID (average of 3 runs) on \texttt{ImageNet-64} (left) and \texttt{FFHQ} (right) when evaluated with Restart sampler. Error bars indicate standard deviations. The $x$-axis indicates the \textit{effective NFE}, which considers all compute costs including the forward and backward passes of the discriminator. \textsc{D-Guidance} has 1.55$\times$ effective NFE compared to the original NFE, whereas our method only has 1.02$\times$ effective NFE compared to the original NFE. The results with original NFE are shown in Figure \ref{fig:benchmark-fid}.}
    \label{fig:restart_fid_effectivenfe}
\end{figure}

\subsection{Additional Results with ADM and VP Diffusion Models}
\label{append:benchmark-admvp-results}
We also experiment with pre-trained diffusion models in \citet{dhariwal2021diffusion} and \citet{song2021scorebased} on \texttt{ImageNet-64} and \texttt{FFHQ} respectively. The results are shown in Figure \ref{fig:adm-vp-results}, where both NFE and effective NFE introduced in Section \ref{append:benchmark-effectivenfe} are considered. As can be observed, our method outperforms baselines across various diffusion models.

\begin{figure}
  \centering
  \begin{subfigure}{\linewidth}
    \includegraphics[width=\linewidth]{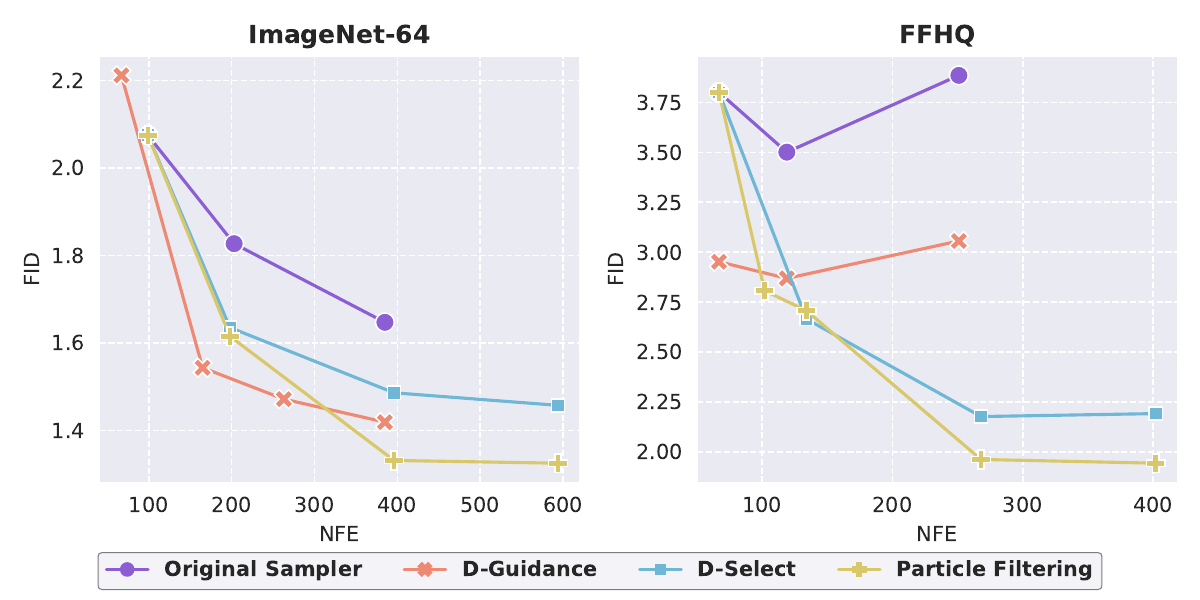}
    \vspace{-5mm}
    \caption{FID against NFE.}
    \label{fig:adm-vp-nfe}
  \end{subfigure}
  \begin{subfigure}{\linewidth}
    \includegraphics[width=\linewidth]{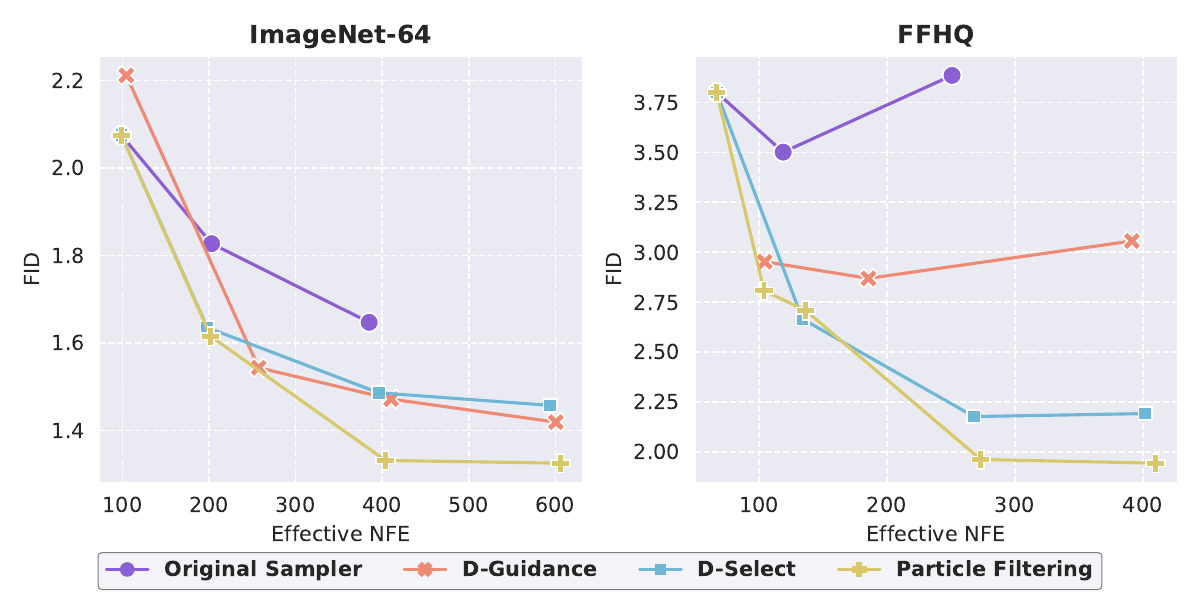}
    \vspace{-5mm}
    \caption{FID against effective NFE.}
    \label{fig:adm-vp-effectivenfe}
  \end{subfigure}
  \vspace{-5mm}
  \caption{FID on \texttt{ImageNet-64} with ADM diffusion model \cite{dhariwal2021diffusion} (left) and \texttt{FFHQ} with VP diffusion model \cite{song2021scorebased} (right) when evaluated with Restart sampler.}
  \label{fig:adm-vp-results}
\end{figure}

\section{Additional Ablation Study}
\label{append:ablation}

In this section, we present additional ablation studies that investigate three design choices in our particle filtering framework. Particularly, we will study the impacts of when the resampling is inserted (\emph{i.e.,} before or after adding noise), the amount of noise added, and the NFE used to generate a single image. We will study their impacts in class-conditioned generation on \texttt{ImageNet-64} and when Restart sampler is used.

\begin{table}
    \centering
    \resizebox{0.42\linewidth}{!}{
    \begin{tabular}{lcc}
    \topline
            &  \textbf{Before} &  \textbf{After} \\
    \midtopline
    $K=2$ & 1.37 & 1.53 \\
    $K=4$ & 1.11 & 1.31 \\
    $K=6$ & 1.07 & 1.23 \\
    \bottomline
    \end{tabular}
    }
    \caption{FID on \texttt{ImageNet-64} when resampling is inserted before or after adding noise.}
    \vspace{-2mm}
    \label{tab:abl-before-noise}
\end{table}

We start by studying \textbf{when is a proper time to insert resampling} during the generation. Particularly, we compare two time points to insert resampling. \ding{182} Resample before adding noise; \ding{183} Resample after adding noise. Table \ref{tab:abl-before-noise} shows the FID for both cases. It can be observed that resampling before adding noise significantly outperforms the counterpart. There could be two potential reasons for the performance difference. First, resampling after adding noise means resampling weights are estimated at higher noise scales, which is less accurate. Second, compared to \ding{183} that adopts a selection-after-exploration procedure, \ding{182} selects particles before exploration, which allows more particles to be sampled around the promising particles with large weights.

\begin{table}
    \centering
    \resizebox{0.65\linewidth}{!}{
    \begin{tabular}{lcc}
    \topline
            &  \textbf{Original noise} &  \textbf{Reduced noise} \\
    \midtopline
    $K=2$ & 1.37 & 1.54 \\
    $K=4$ & 1.11 & 1.37 \\
    $K=6$ & 1.07 & 1.31 \\
    \bottomline
    \end{tabular}
    }
    \caption{FID on \texttt{ImageNet-64} with original and reduced noise.}
    \label{tab:abl-noise-amount}
    \vspace*{-3mm}
\end{table}

Following the above intuition, we further explore the impact of the \textbf{amount of noise added}. Specifically, we change the restart configuration such that the variance of the added noise at time \e{t} is reduced to \e{t^2}, while keeping all other configurations the same. Table \ref{tab:abl-noise-amount} shows the FID when the original and reduced noise is added.
The results indicate the importance of a large added noise.

\begin{figure}
    \centering
    \includegraphics[width=0.9\linewidth]{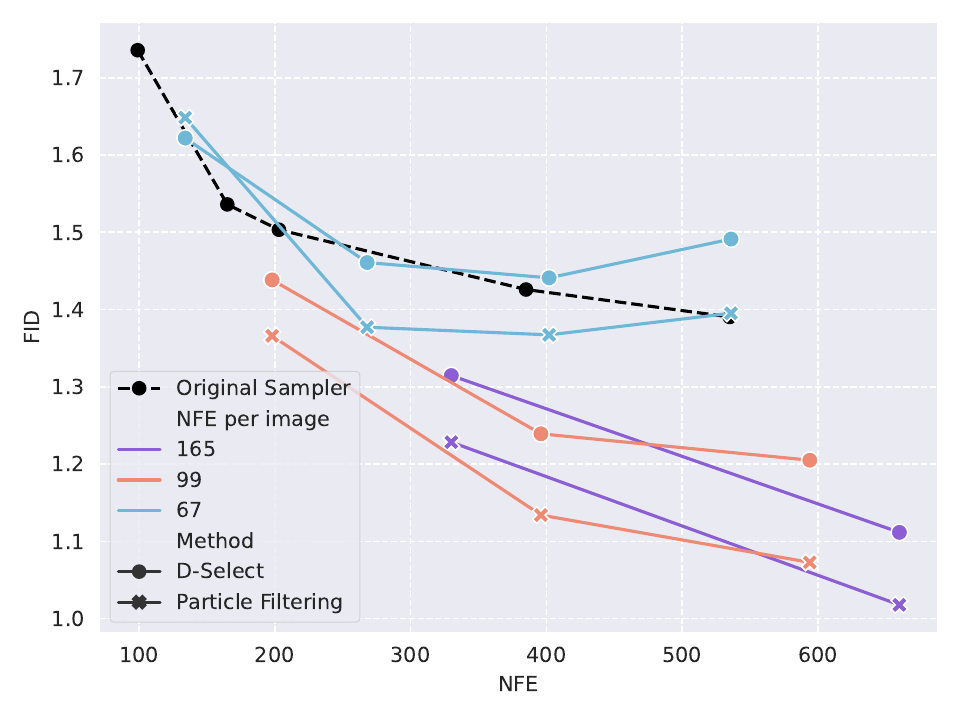}
    \caption{FID on \texttt{ImageNet-64} when \textsc{D-select} and our \textsc{PF} method are applied to different NFEs per image.}
    \vspace{-2mm}
    \label{fig:abl-discretization}
\end{figure}
Finally, we explore the impact of the \textbf{NFE for each image}. Instead of fixing an NFE and only varying the number of particles, we now evaluate our method when it is applied to different NFEs. Figure \ref{fig:abl-discretization} shows the FID when each image is generated with \e{\text{NFE}=67, 99, 165}. For each NFE, we further report the performance when \e{K=2,4,6} images are generated (except for \e{\text{NFE}=165} due to the large cost). There are two observations from the figure. First, \textsc{PF} consistently outperforms \textsc{D-select} across all NFEs, again demonstrating its advantages. Second, different NFEs present varying performance trends. When the number of denoising steps is small (\e{\text{NFE}=67}), FID reaches a plateau at \e{K=4} particles, and further increasing the value of \e{K} does not improve performance due to the large discretization error. On the contrary, FID continues decreasing as \e{K} increases when the discretization error is small (\e{\text{NFE}=99,165}). Particularly, generating \e{K=4} particles when \e{\text{NFE}=165} achieves the state-of-the-art FID of 1.02 on \texttt{ImageNet-64}.

\begin{figure*}
    \centering
    \includegraphics[width=0.85\linewidth]{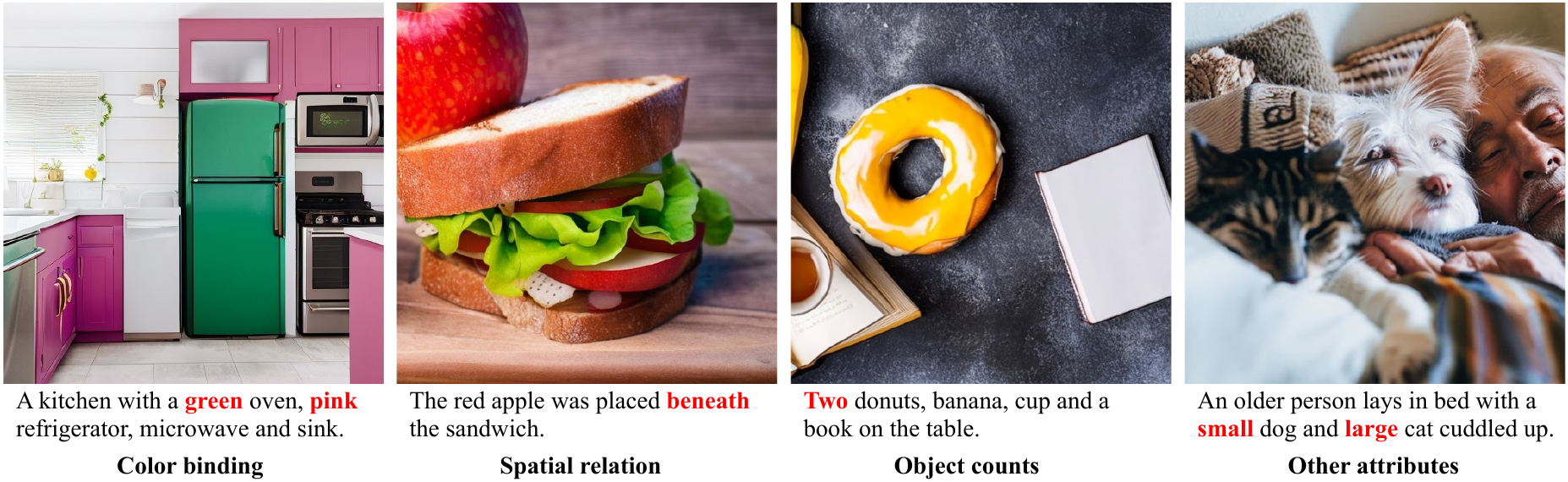}
    \caption{Categories of common failure cases of our method. Errors are highlighted in \textcolor{red}{red}.}
    \vspace{-2mm}
    \label{fig:failure-case}
\end{figure*}

\section{Generated Samples}
\label{append:image-sample}
In this section, we present more generated samples for our method and baselines.

Figure \ref{fig:failure-case} illustrates four types of failure cases of our method. In general, our method tends to miss the attributes that are not captured by the object detector, such as colors, locations, counts, sizes, \emph{etc}, as shown by the four images, respectively. This is because our framework drops the correction term for object characteristics in Eq. \eqref{eq:weight_decomp}. However, our method remains flexible to incorporate other desired measures into the particle weight calculation.

Figures \ref{fig:imagenet-samples} and \ref{fig:ffhq-samples} show the uncurated samples on \texttt{ImageNet-64} and \texttt{FFHQ} respectively. As can be observed, both \textsc{D-guidance} and \textsc{PF} generate images with higher quality than the original sampler. Compared to \textsc{D-guidance}, our method generates images with closer and more detailed views (\emph{e.g.,} class ``mongoose'').

\begin{figure*}
    \centering
    \includegraphics[width=0.88\linewidth]{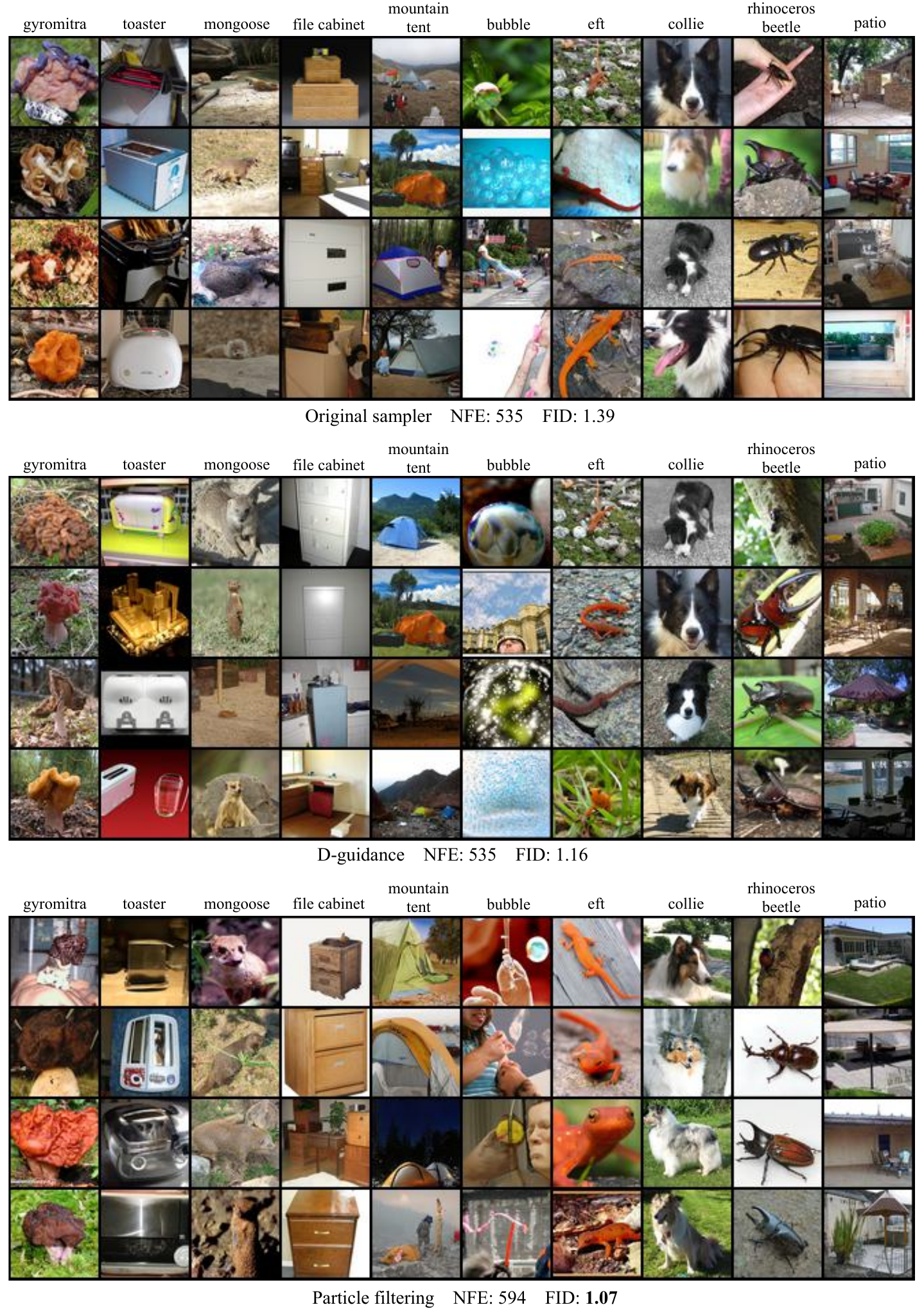}
    \vspace{-2mm}
    \caption{Uncurated samples on \texttt{ImageNet-64} for our method and baselines. All methods use Restart sampler.}
    \label{fig:imagenet-samples}
\end{figure*}

\begin{figure*}
    \centering
    \includegraphics[width=0.88\linewidth]{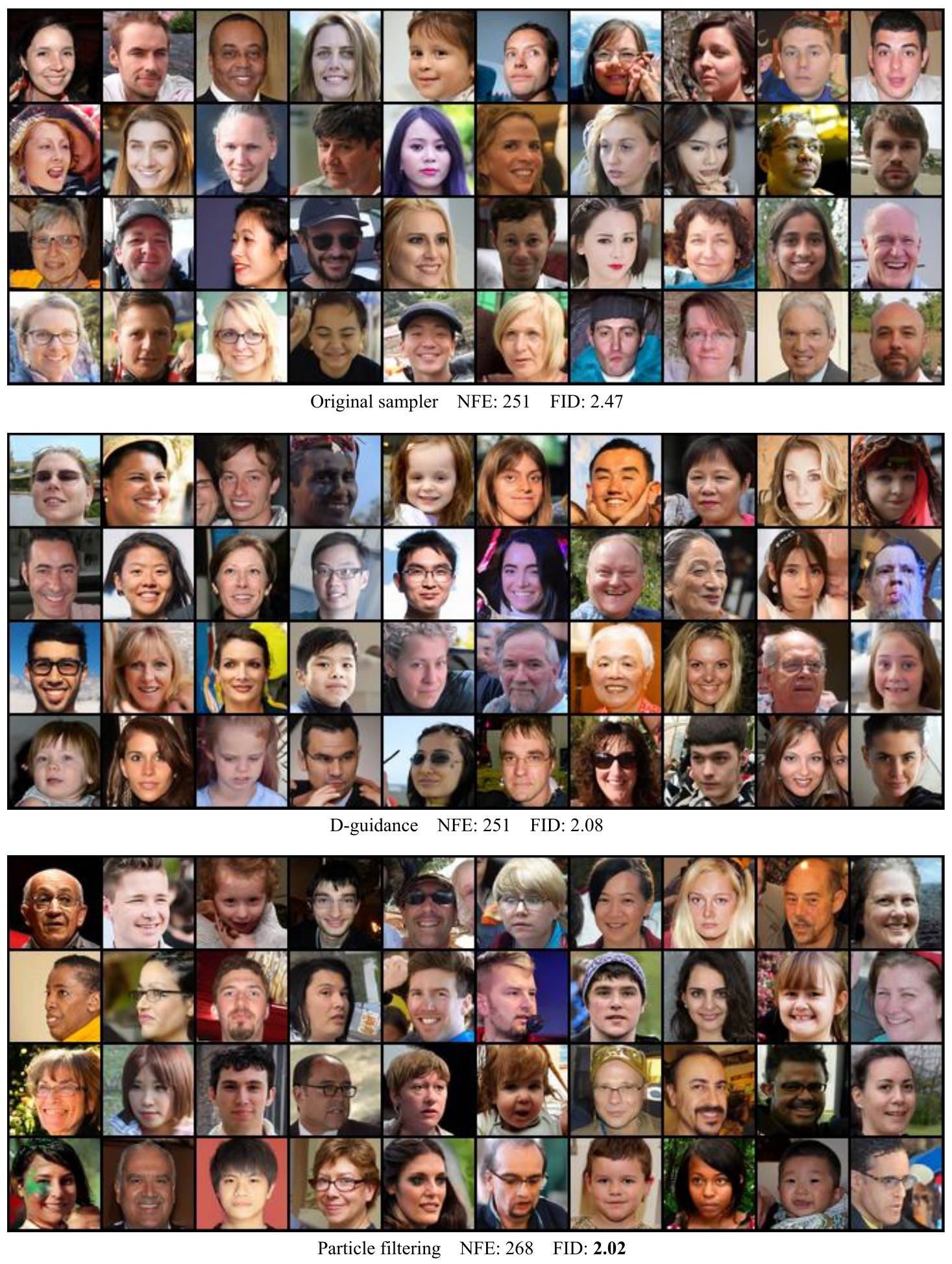}
    \vspace{-2mm}
    \caption{Uncurated samples on \texttt{FFHQ} for our method and baselines. All methods use Restart sampler.}
    \label{fig:ffhq-samples}
\end{figure*}

\section{Details of Subjective Evaluation}
We recruit human annotators to evaluate both object occurrence and image quality of generated images. For object occurrence, annotators are provided with an image and a list of objects, and they are asked to identify whether each object occurs in the image or not. For image quality, annotators are given two images generated based on the same caption, where one image is generated by \textsc{PF-hybrid} and the other image by a baseline. The caption is not revealed to annotators in order to make the evaluation focus more on image quality instead of image and text alignment. The annotation interfaces for object occurrence and image quality are shown in Tables \ref{tab:object_instruction} and \ref{tab:quality_instruction} respectively.

\begin{table*}[t]
	\begin{tabular}{p{1\linewidth}}
	\topline
    \textbf{Instructions:}\\
    \midrule
You will be given an image generated by some AI algorithm, your task is to \textbf{identify all objects} that appear in the image.\\

Notes:\\
\quad\textbullet \, You should \textbf{ignore the characteristics} of the object such as color and count when deciding whether it appears or not. E.g., if the object is ``a red apple,'' you should still select it even if the image shows a green apple.\\
\quad\textbullet \, Only select objects that you are \textbf{confident} about. \emph{E.g.}, the ``knife'' in the image below is not clear, so you should not select it.\\
    \vspace{0.05in}
    \textbf{Example}: \\

    \vspace{-5mm}
    \begin{center}
    \includegraphics[width=0.3\linewidth]{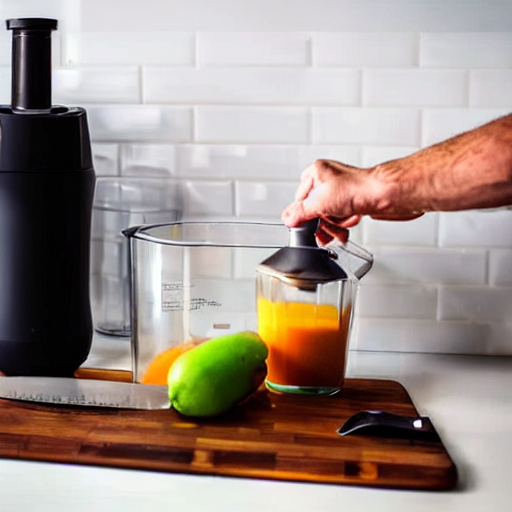}
    \end{center}
    
    Do the following objects appear in the image? \\
    \textbullet \, A black blender: Yes \quad \textbullet \, bottle: No \quad \textbullet \, glass: Yes \quad \textbullet \, cutting board: Yes \quad \textbullet \, knife: No \\
    \\
    
	\midtopline
    \textbf{Task:} \\
    \midrule
    Please identify \textbf{all objects} appear in the image \\
    \vspace{-3mm}
    \begin{center}
    \includegraphics[width=0.3\textwidth]{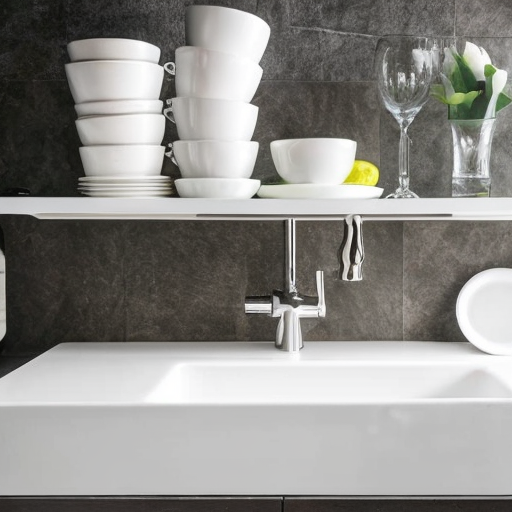}
    \end{center}
 
     $\square$ white kitchen sink \quad $\square$ white coffee cups \quad $\square$ wine glasses \quad $\square$ knives \quad $\square$ None of the above \\
     \bottomline
	\end{tabular}
    \caption{Instructions and an example task for the subjective evaluation on object occurrence.}
    \label{tab:object_instruction}
\end{table*}

\begin{table*}[t]
	\begin{tabular}{p{1\linewidth}}
	\topline
    \textbf{Instructions:}\\
    \midrule
In this task, you will see two images that are generated based on a text caption by different AI algorithms. You will be asked to evaluate \textbf{which image looks more real and natural}.\\

    \vspace{0.05in}
    \textbf{Example}: \\

    \vspace{-5mm}
    \begin{center}
    
    \includegraphics[width=0.6\linewidth]{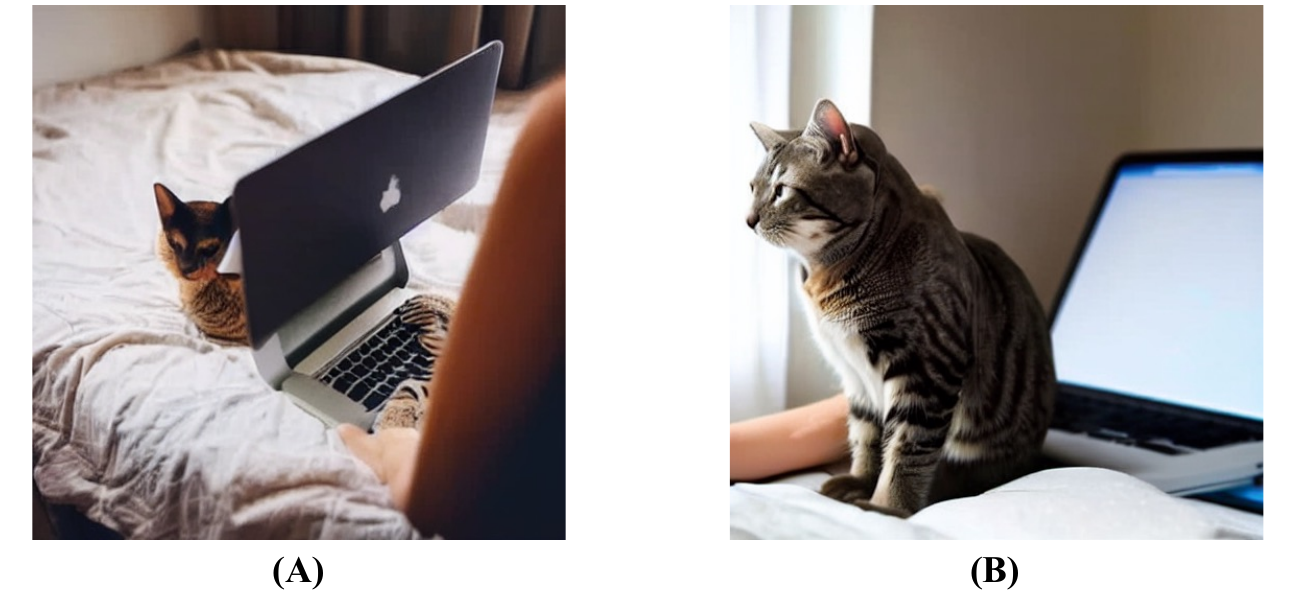}
  
    \end{center}

    Image (B) generates a natural cat and laptop, whereas the laptop in image (A) is distorted, and the human body is also unnatural. Therefore, \textbf{image (B) is better}. \\
    \\
    
	\midtopline
    \textbf{Task:} \\
    \midrule
    Which image looks more natural and like a real photo? \\
    \vspace{-3mm}
    \begin{center}
    \includegraphics[width=0.61\textwidth]{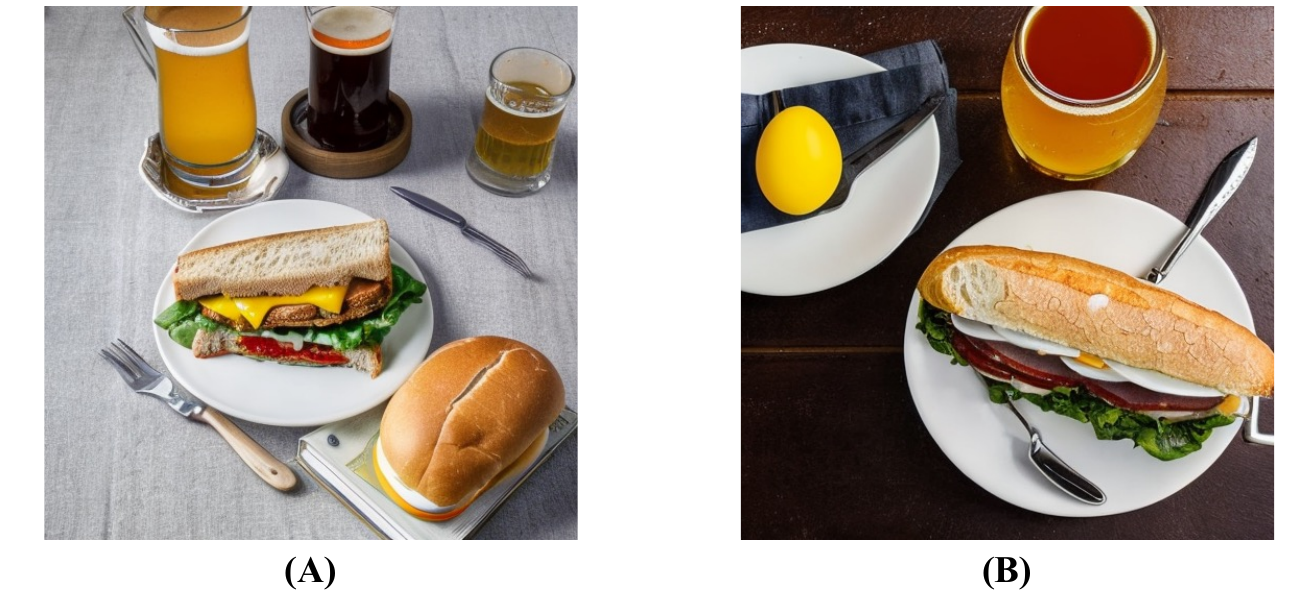}
    \end{center}
 
     $\square$ (A) is better \quad $\square$ (B) is better \\
     \bottomline
	\end{tabular}
    \caption{Instructions and an example task for the subjective evaluation on image quality.}
    \vspace{-10mm}
    \label{tab:quality_instruction}
\end{table*}

\end{alphasection}